\documentclass[lettersize,journal]{IEEEtran}
\usepackage{amsmath,amsfonts}
\usepackage{algorithmicx}
\usepackage{algorithm}
\usepackage{array}
\usepackage[caption=false,font=normalsize,labelfont=sf,textfont=sf]{subfig}
\usepackage{textcomp}
\usepackage{stfloats}
\usepackage{url}
\usepackage{verbatim}
\usepackage{graphicx}
\usepackage{cite}
\usepackage{amssymb}
\usepackage[table,dvipsnames]{xcolor}
\usepackage{multirow}
\usepackage{stackengine}
\usepackage{siunitx}
\usepackage[framemethod=TikZ]{mdframed}
\usepackage{tikz}
\usepackage{schemabloc}
\usepackage[hidelinks]{hyperref}
\usepackage{acro}
\usepackage{float}
\usepackage{algpseudocode}
\usepackage{svg}
\usepackage{orcidlink}

\usetikzlibrary{trees, calc, shadings, shapes.gates.logic.US, circuits.logic.US, positioning, arrows, shapes.geometric, shapes.symbols, fit, shadows, circuits, angles, quotes, arrows.meta, decorations.pathreplacing, trees}

\definecolor{renault}{HTML}{F8B700}
\definecolor{gps_color}{HTML}{17BECF}
\definecolor{speed_color}{HTML}{F14CC1}
\definecolor{safetycolor}{HTML}{00b200}

\newcolumntype{K}[1]{>{\centering\arraybackslash}m{#1}}

\makeatletter
\newcommand{\labeltext}[2]{%
  \@bsphack
  \csname phantomsection\endcsname 
  \def\@currentlabel{#1}{\label{#2}}%
  \@esphack
}
\makeatother

\DeclareAcronym{av}{
  short = AV,
  long  = autonomous vehicle
}
\DeclareAcronym{fmp}{
  short = FMP,
  long  = field monitoring process
}
\DeclareAcronym{hara}{
  short = HARA,
  long  = Hazard Analysis and Risk Assessment
}
\DeclareAcronym{fusa}{
  short = FuSa,
  long  = functional safety
}
\DeclareAcronym{asil}{
  short = ASIL,
  long  = automotive safety integrity level
}
\DeclareAcronym{sotif}{
  short = SOTIF,
  long  = safety of the intended functionality
}
\DeclareAcronym{aspice}{
  short = ASPICE,
  long  = automotive software process capability determination
}
\DeclareAcronym{odd}{
  short = ODD,
  long  = Operational Design Domain
}
\DeclareAcronym{stpa}{
  short = STPA,
  long  = Systems-Theoretic Process Analysis
}
\DeclareAcronym{hazop}{
  short = HAZOP,
  long  = Hazard and Operability Study
}
\DeclareAcronym{fsc}{
  short = FSC,
  long  = Functional Safety Concept
}
\DeclareAcronym{tsc}{
  short = TSC,
  long  = Technical Safety Concept
}
\DeclareAcronym{fsr}{
  short = FSR,
  long  = Functional Safety Requirement
}
\DeclareAcronym{tsr}{
  short = TSR,
  long  = Technical Safety Requirement
}
\DeclareAcronym{sg}{
  short = SG,
  long  = safety goal
}
\DeclareAcronym{ss}{
  short = SS,
  long  = safety state
}
\DeclareAcronym{os}{
  short = OS,
  long  = operating scenario
}
\DeclareAcronym{fmea}{
  short = FMEA,
  long  = failure mode and effect analysis
}
\DeclareAcronym{fta}{
  short = FTA,
  long  = Fault Tree Analysis
}
\DeclareAcronym{sw}{
  short = SW,
  long  = software
}
\DeclareAcronym{hw}{
  short = HW,
  long  = hardware
}
\DeclareAcronym{ai}{
  short = AI,
  long  = Artificial Intelligence
}
\DeclareAcronym{bt}{
  short = BT,
  long  = behavior tree
}
\DeclareAcronym{npc}{
  short = NPC,
  long  = non-player character
}
\DeclareAcronym{fsm}{
  short = FSM,
  long  = finite-state machine
}
\DeclareAcronym{ros}{
  short = ROS,
  long  = Robot Operating System
}
\DeclareAcronym{gui}{
  short = GUI,
  long  = graphical user interface
}

\tikzstyle{root}=[isosceles triangle, isosceles triangle apex angle=100, shape border rotate=270, draw,fill=black!0, minimum size=5mm]
\tikzstyle{fallback}=[rectangle,draw,fill=black!10, minimum size=8mm]
\tikzstyle{sequence}=[rectangle,draw,fill=black!20, minimum size=8mm]
\tikzstyle{state}=[rectangle,draw,fill=safetycolor!30, minimum size=8mm]
\tikzstyle{condition}=[circle,draw,fill=yellow!50, minimum size=8mm]
\tikzstyle{decorator}=[diamond,draw,fill=red!30, minimum size=8mm]
\tikzstyle{action}=[rectangle,draw,fill=renault!60, minimum size=8mm, rounded corners=2mm]
\tikzstyle{parallel}=[rectangle,draw,fill=black!30, minimum size=8mm]
\tikzstyle{subtree}=[rectangle,draw,fill=blue!20, minimum size=8mm, rounded corners=2mm]

\tikzstyle{block_iso26262}=[rectangle, draw, fill=renault!10, text width=6em, text centered, rounded corners, minimum height=3em, node distance=0.4cm]
\tikzstyle{line}=[draw, -latex']
\tikzstyle{line_double}=[draw, double]
\tikzstyle{line_v}=[draw,-{Fast Triangle[color=black]},line width=3pt]
\tikzstyle{line2}=[draw, latex'-latex']
\tikzstyle{block_aspice}=[rectangle, draw, fill=safetycolor!30, text width=8em, text centered, rounded corners, minimum height=4em, node distance=0.5cm]
\tikzstyle{block_sotif}=[rectangle, draw, fill=renault!30, text width=6em, text centered, rounded corners, minimum height=3em, node distance=0.5cm]
\tikzstyle{curly}=[brace, draw, amplitude=10pt, raise=1mm, mirror, aspect=0.5]

\begin{document}

\title{Behavior Trees in Functional Safety \\ Supervisors for Autonomous Vehicles}

\author{Carlos Conejo$^{\orcidlink{0009-0007-7233-5650}}$, Vicenç Puig$^{\orcidlink{0000-0002-6364-6429}}$, Bernardo Morcego$^{\orcidlink{0000-0002-6944-7519}}$, Francisco Navas$^{\orcidlink{https://orcid.org/0000-0002-6066-0612}}$,  Vicente Milanés$^{\orcidlink{https://orcid.org/0000-0001-7096-6925}}$ 
    \thanks{Research partially funded by the Spanish State Research Agency (AEI) and the European Regional Development Fund (ERFD) through the SaCoAV project (ref. PID2020-114244RB-I00). Also funded by Renault through the Industrial Doctorate "Safety of Autonomous Vehicles" (ref. C12507). }
    \thanks{Carlos Conejo and Vicenç Puig are with Institut de Robòtica i Informàtica Industrial (CSIC-UPC), Universitat Politècnica de Catalunya-BarcelonaTech, Barcelona, Spain.
            {\tt\small Email: \{carlos.conejo, vicenc.puig\}@upc.edu}}%
    \thanks{Carlos Conejo, Vicenç Puig and Bernardo Morcego are with the Automatic Control Group (CS2AC-UPC), Universitat Politècnica de Catalunya-BarcelonaTech, Terrassa, Spain.
            {\tt\small Email: bernardo.morcego@upc.edu}}%
    \thanks{Carlos Conejo, Francisco Navas and Vicente Milanés are with the Renault Group, Spain.
            {\tt\small Email: \{francisco-martin.navas, vicente.milanes\}@renault.com}}%
}

\markboth{ IEEE TRANSACTIONS ON INTELLIGENT TRANSPORTATION SYSTEMS}%
{Shell \MakeLowercase{\textit{et al.}}: A Sample Article Using IEEEtran.cls for IEEE Journals}

\IEEEpubid{\copyright~2024 IEEE. }

\maketitle

\begin{abstract}

The rapid advancements in \acl{av} \acl{sw} present both opportunities and challenges, especially in enhancing road safety. The primary objective of \aclp{av} is to reduce accident rates through improved safety measures. However, the integration of new algorithms into the \acl{av}, such as \acl{ai} methods, raises concerns about the compliance with established safety regulations. This paper introduces a novel \acl{sw} architecture based on \aclp{bt}, aligned with established standards and designed to supervise vehicle \acl{fusa} in real time. It specifically addresses the integration of algorithms into industrial road vehicles, adhering to the ISO~26262. 

The proposed supervision methodology involves the detection of hazards and compliance with functional and technical safety requirements when a hazard arises. This methodology, implemented in this study in a Renault Mégane (currently at SAE level 3 of automation), not only guarantees compliance with safety standards, but also paves the way for safer and more reliable autonomous driving technologies.

\end{abstract}

\acresetall


\section{Introduction}
    \subsection{Motivation}
        \Acp{av} aim to make roads safer by reducing accidents caused by human errors \cite{fagnant2015preparing}. Unlike regular cars, \acp{av} use advanced technology to do the thinking, seeing and moving that a human driver would normally do. The \ac{hw} of \acp{av} functions like the human senses. It uses sensors to collect information about the surroundings and actuators to control the steering wheel and acceleration pedal, similar to the human motor system. Meanwhile, \ac{sw} acts as the human brain. It uses algorithms to make driving decisions based on data collected from sensors.

        Ensuring safety in the process of replacing human control with technology in \acp{av} is crucial. This involves verifying that sensors accurately collect data, algorithms correctly interpret this information and create a safe navigation path, and actuators effectively execute the given commands. Due to the complex and interdependent nature of safety in \acp{av} \cite{koopman2017autonomous}, adherence to established safety standards is essential. Commercial self-driving cars must meet the \ac{sw} and \ac{hw} requirements of ISO~26262 and ISO~21448 to ensure this level of safety.
        
        This article narrows its scope within the range of approaches to meeting safety standards from a scientific perspective. It focuses on ensuring runtime compliance with \acf{fusa}, addressed in ISO~26262. With the ongoing evolution of \ac{av} technology, specifically at SAE levels 4+ \cite{iso22736}, \ac{sw} updates are developed and applied regularly. This research aims to create a safety supervision framework that verifies, in real time, the compliance of the navigation \ac{sw} with the \ac{tsc}, defined before product development, according to ISO~26262. 
        
    \subsection{Related Work}
        Current industrial regulations and the state-of-the-art in the area of safety and supervision in autonomous driving are summarized in this section.
        
        \subsubsection{Safety Regulations for \acp{av}}
            This subsection provides a concise overview of existing industry standards. They aim to establish consensus among various companies and set baseline requirements. These standards are designed to improve safety in \ac{av} by reducing the risks and potential harms associated with existing and emerging systems.

            The initial step in adhering to standards involves detailing the vehicle's operational conditions, encapsulated by the concept of \acf{odd}. This includes elements such as speed limits, anticipated weather conditions, and potential interactions with various obstacles. Establishing \ac{odd} is crucial as it lays the foundation for the definition of the system requirements for the \ac{fusa} standard, ISO~26262.

            \IEEEpubidadjcol

            ISO~26262 \cite{iso26262} is a standard dedicated to the \ac{fusa} of electrical and/or electronic systems in production automobiles. It emphasizes ensuring safe operation throughout the lifecycle of automotive equipment. It plays an important role, especially as vehicles increasingly depend on sophisticated electronic components and \ac{sw} for essential functions such as braking, steering, and acceleration. This standard is required for automotive manufacturers and suppliers in the design and implementation of safe vehicle systems. It deals with the mitigation of risk from system failures. On the other hand, the \ac{sotif} standard, ISO~21448 \cite{sotif}, provides guidance on design, verification and validation focusing on mitigating, preventing, and controlling risks arising from system failures. This standard assesses whether the required safety functionalities can be ensured in unforeseen conditions without failure. It also takes into account external factors, such as weather conditions and potential misuse, to ensure comprehensive safety coverage.
            
            After defining the \ac{av} functional requirements (item definition) and the \ac{odd}, the \ac{hara} is carried out. This involves the identification of hazards and the assessment of risk, which guide the determination of \acp{sg} based on severity, controllability and exposure, summarized in an \ac{asil}. These \acp{sg} are essential to maintain safety and mitigate the risks of malfunctioning electrical and electronic systems. Techniques such as \ac{stpa}, originally for aerospace\cite{4526677}, and \ac{hazop} support the systematic approach in \ac{hara} \cite{abdulkhaleq2017using}. Incorporating \ac{sotif} into ISO~26262 \ac{hara} enables a more comprehensive safety analysis, enhancing the safety of \acp{av} \cite{schildbach2018}.

            Following ISO~26262 guidelines, the \ac{fsc} is established, providing a detailed overview of the system to achieve \acp{sg}. Developed from \ac{hara} outcomes, it outlines necessary functions for the maintenance of safety. The high-level architecture of the system is subjected to an iterative analysis to evaluate its impact on \acp{sg} from \ac{hara} \cite{graubohm2019}. \Acp{fsr} specify methods for diagnosing and mitigating faults that threaten these \acp{sg}. To confirm \ac{fsc} compliance with \acp{sg}, safety analyses like qualitative \ac{fmea} and \ac{fta} are conducted. 

            Once the concept phase has ended, the product development begins with the \acf{tsc}, which translates the \ac{fsc} into system-level implementation. This phase involves defining \acp{ss}, specific operational conditions or modes that aim to achieve or maintain \acp{sg}, particularly under identified risks. \Acp{ss} encompass activating mitigation actions or fallback procedures for an appropriate system response to events or faults. Recent studies on \ac{tsc} evaluation demonstrate handling functional safety across various \ac{av} levels using the \ac{stpa} methodology \cite{abdulkhaleq2017systematic}.

            SAE levels 4+ \acp{av}, representing fully autonomous navigation, often rely on non-deterministic, \ac{ai}-based deep learning algorithms, complicating explainability and repeatability, a crucial aspect of ISO~26262 and \ac{sotif}. Currently, there is no specific standard dedicated to ensuring the safety of SAE levels 4+ \acp{av}. Therefore, researchers have explored various methodologies within the ISO~26262 guidelines during different stages of the product lifecycle \cite{8479057}. Another research direction is the development of a new standard, UL~4600 \cite{koopman2019}, primarily applicable in the USA. This standard, inspired by agile and iterative aviation safety approaches, emphasizes continuous evaluation and improvement of residual risk in \ac{av} systems.

            In automotive \ac{sw} development, \Ac{aspice} plays a crucial role, acting as a key framework to improve the quality of the process. It offers a structured methodology for \ac{sw} engineering, enhancing the development life cycle. Implementing \ac{aspice}'s guidelines, based on the V-model, enables organizations to reach higher maturity levels in their development processes. This not only aligns with ISO~26262 compliance, but also increases \ac{sw} quality, indirectly improving the safety integrity of automotive products. Practical applications of \ac{aspice} in an \acp{av} demonstrate its efficacy \cite{macher2017}.
            
            Recent proposals for architectures that guarantee \ac{fusa} \cite{ulbrich2017functional}, \cite{bagschik2018system} serve as foundational references for designing an \ac{av} supervisor. In the mentioned articles, the adhesion to the ISO~26262 requirements is performed due to modularity, granularity, simplicity, hierarchical design, maintainability and testability. The development of this framework's C++ code extends the ISO~26262 and \ac{aspice} \ac{sw} guidelines \cite{tabani2019}.

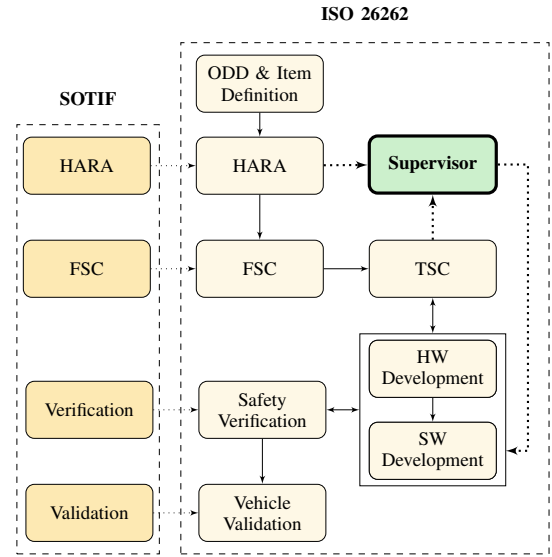
\begin{figure}[ht]
                \small
                \centering       
                \resizebox{0.8\columnwidth}{!}{\begin{tikzpicture}
                
                  \node [block_iso26262] (item) {\acs{odd} \& Item \\ Definition};
                  \node [block_iso26262, below=of item] (hara) {\acs{hara}};
                  \node [block_iso26262, below=0.75cm of hara] (fsc) {\acs{fsc}};
                
                  \node [block_iso26262, right=0.75cm of fsc] (tsc) {\acs{tsc}};
                  \node [block_iso26262, below=0.7cm of tsc] (hw) {HW};
                  \node [block_iso26262, below=of hw] (sw) {SW};
                  \node [draw=black, inner xsep=0.15cm, inner ysep=0.1cm, fit=(hw) (sw)] (pd) {};
                  \node [block_iso26262, below=0.7cm of tsc] (hw) {HW \\ Development};
                  \node [block_iso26262, below=of hw] (sw) {SW \\ Development};
                  \node [above=0.25cm of pd.west] {};  
                  \node [block_iso26262, left=0.55cm of pd] (safetyVer) {Safety \\ Verification};
                  \node [block_iso26262, below=0.75cm of safetyVer] (vehicleVal) {Vehicle Validation};
                  \node [block_iso26262, above=0.75cm of tsc, fill=safetycolor!20, node distance=3cm, line width=1.5pt] (bt) {\textbf{Supervisor}};
                  
                  \node [block_sotif, left=0.75cm of hara] (sotif1) {\acs{hara}};
                  \node [block_sotif, left=0.75cm of fsc] (sotif2) {\acs{fsc}};
                  \node [block_sotif, left=0.75cm of safetyVer] (sotif3) {Verification};
                  \node [block_sotif, left=0.75cm of vehicleVal] (sotif4) {Validation};

                  \node [draw=black, dashed, inner xsep=0.1cm, inner ysep=0.2cm, fit=(sotif1) (sotif2) (sotif3) (sotif4)] (sotif) {};
                  \node [above=0.2cm of sotif.north] {\textbf{\acs{sotif}}}; 
                  \coordinate (East) at ([xshift=0.6cm]bt.east);
                  \coordinate (East2) at ([xshift=0.15cm]sw.east);
                  \node [draw=black, dashed, inner xsep=0.25cm, inner ysep=0.2cm, fit=(item) (hara) (fsc) (tsc) (hw) (sw) (safetyVer) (vehicleVal) (East)] (iso26262) {};
                  \node [above=0.25cm of iso26262.north] {\textbf{ISO~26262}};   
                
                  \path [line] (item) -- (hara);
                  \path [line] (hara) -- (fsc);
                  \path [line] (fsc.east) -- (tsc.west);
                  \path [line2] (tsc) -- (pd);
                  \path [line] (hw) -- (sw);
                  \path [line2] (pd) -- (safetyVer);
                  \path [line] (safetyVer) -- (vehicleVal);

                  \path [line, dotted] (sotif1) -- (hara);
                  \path [line, dotted] (sotif2) -- (fsc);
                  \path [line, dotted] (sotif3) -- (safetyVer);
                  \path [line, dotted] (sotif4) -- (vehicleVal);
                  \path [line, dotted, line width=1pt] (hara) -- (bt);
                  \path [line, dotted, line width=1pt] (tsc) -- (bt);
                  \path [line, dotted, line width=1pt] (bt.east) -- ++ (0.5,0) |- (East2.east);
                  
                \end{tikzpicture}}
                \caption{\footnotesize Industrial procedure for the design of self-driving car's \ac{sw}. The block diagram is divided into two groups, each corresponding to the safety standard followed. The \textit{Supervisor} block designed in the article is part of the \ac{sw} development and depends on the \ac{hara} and \ac{tsc}.}
                \label{fig:standards}
            \end{figure}

            Fig.~\ref{fig:standards} shows the interaction of the safety standards, graphically defining the scope of this article, represented by the \textit{Supervisor} block. 

        \subsubsection{Safety \ac{sw} for \acp{av}}
            In \ac{av} navigation \ac{sw}, various safety-handling approaches are used. The first method involves integrating safety parameters into low-level algorithms to reduce risk \cite{8772139}. The second approach, also applied to low-level \ac{sw}, is the fusion of data from multiple sources to enhance reliability \cite{ifqir2022fault}, \cite{prieto2020localization}, \cite{vicens2022localization}. The third strategy, relevant to the high-level \ac{sw}, focuses on designing fault-tolerant control architectures \cite{blanke2006}. The combination of these techniques is not only feasible but also required by safety regulations to ensure complete risk mitigation in \ac{av} systems. 
            
            Systematic errors in \ac{sw} typically arise from incorrect requirements specifications or coding bugs. Unlike \ac{hw}, achieving reliable redundancy in \ac{sw} is challenging, as different versions created by separate teams often replicate similar errors. Consequently, the consensus in the field is that the development process is the most critical aspect of \ac{sw} quality. Adherence to the \ac{aspice} framework and its best practices is essential to mitigate these systematic errors and ensure high-quality software development \cite{schildbach2018}.
            
            At the beginning of this century, the applications of autonomous systems for fault diagnosis, fault-tolerant control, and control reconfiguration were introduced \cite{blanke2006}. This approach proposes an execution level and a supervision level to ensure safety and is still prevalent in \acp{av}. 
            
            Hybrid automatons, which combine discrete events with continuous variables, have been widely used to create supervisors \cite{lemmon1999supervisory}, \cite{vento2013hybrid}, \cite{heffernan2014runtime}. However, in complex high-dimensional systems such as \acp{av}, hybrid automatons face challenges in adapting to new data and scenarios due to their complexity.

            An alternative supervision method involves semi-Markov processes, beneficial for modeling time-dependent behaviors and complex scenarios with variable time intervals. Despite their advantages, semi-Markov processes pose computational challenges for \acp{av}. Furthermore, the calculation of reliability and availability is complex and prone to errors \cite{kaalen2019}, making them less practical for \acp{av} applications.

            The combination of \ac{fusa} and \ac{sotif} standards to design \ac{av} supervisors has been a subject of study \cite{dorff2020}, \cite{stolte2020towards}, \cite{CUER201829}. These supervisors were originally conceptualized to provide safety assurances for each system independently. In contrast, this study seeks to depart from this system-specific approach. The goal is to develop a supervisor architecture that is not only standardized but also adaptable, offering a more universal solution to ensure safety across various \ac{av} systems. This approach aims to create a more uniform and flexible framework for \ac{sw} supervision in \acp{av}. 

        \subsubsection{Behavior Trees}
            
            In recent times, \acp{bt}, initially created to enhance the \ac{ai} of \acp{npc} in video games \cite{isla2008halo}, have found applications in robotics \cite{colledanchise2018behavior}, \cite{iovino2022survey}. These \acp{bt} are capable of representing hybrid automatons \cite{marzinotto2014towards}, \cite{colledanchise2017how}, and offer several advantages over hybrid automatons and semi-Markov processes. These advantages include modularity, which allows developers to build complex behaviors from simpler and reusable components; flexibility, which makes it easy to modify and extend behaviors without affecting the entire system; scalability, making architectures more manageable for large and complex behavior sets; and maintainability, being easier to debug and maintain due to their structured and clear format \cite{ogren2012increasing}. 

            \Acp{bt} feature a hierarchical, tree-like structure that facilitates the creation, modification, and extension of complex behaviors. Widely used for decision-making algorithms in video games and robotics, \acp{bt} are a popular choice \cite{palma2011extending}, \cite{bagnell2012integrated}, \cite{ogren2012increasing}, \cite{marzinotto2014towards}, \cite{olsson2016behavior}, \cite{rovida2018motion}, \cite{paxton2019representing}. 
            
            In practical implementations, some of these \acp{bt} have been deployed using the \ac{ros} and the \textit{BehaviorTree.CPP} library \cite{marzinotto2014towards}, \cite{faconti2018behaviortree}. Additionally, the \textit{Groot} \ac{gui} has been used to edit trees and monitor their performance in real time \cite{tadewos2019decentralized}, \cite{ghzouli2020behavior}. Once \acp{bt} are analyzed, they can potentially become a very suitable technique to supervise the \ac{av} \ac{sw}. This application has been previously proven in the aviation field \cite{lindsay2010safetyassessment}. 
            
        \subsubsection{Safety Testing}
            Testing is a crucial component of safety assurance in autonomous driving, focusing on ensuring that all predefined requirements are met in a finite set of distinct scenarios. The scenario-based approach currently stands as the predominant method for testing in autonomous driving. This approach involves generating traffic situations to evaluate the performance of self-driving cars. These scenarios can be realistic or simulated \cite{Survey_Safety}. Initially, simulations are preferred for safety as they allow examination and refinement in a controlled environment. Afterwards, and depending on resource availability, real-world tests are conducted to validate the vehicle's performance under driving conditions.
            
            On the other hand, falsification looks for singular scenarios in which the model violates these requirements. It reveals safety-critical faults in the system and, using iterative methods, can help improve overall performance \cite{koschi2019computationally}.
    
            Finally, formal verification provides proof of correctness but lacks scalability to complex systems. A combination of a scenario-based approach and verification is proposed to be an efficient approach to the safety of \acp{av} \cite{Survey_Safety}. Scalability challenges are faced in Mobileye's research \cite{shalevshwartz2018formal}.

            At the system's \ac{sw} level, the accuracy of training and validation data is critical to safety because it reduces hazards, such as unintended biases. The importance of acquiring accident data to learn how to avoid accidents in the most critical scenarios of autonomous driving is high \cite{winkle2016safety}. 

    \subsection{Contributions}

        The main objective of this paper is to present a systematic methodology for designing supervisors for SAE level 4+ \acp{av}, aligning with established safety standards. The contribution lies in four different aspects:

        \begin{itemize}
            \item Formalization of analogies between static and dynamic (run-time) safety analyses.
            \item Automatization of the design of a \ac{fmp} from the ISO~26262 \ac{fusa} standard. 
            \item Development of a \ac{fusa} supervisor, responsible for recovering the system if a fault occurs, based on static safety analysis. 
            \item Implementation of the methodology in an autonomous Renault Mégane.
        \end{itemize}

        The construction of the supervisor, based on ISO~26262, consists of creating a new \ac{sw} supervisor architecture, based on \acp{bt}. This architecture ensures compliance with all \acp{sg} and facilitates the \ac{av}'s transition to the appropriate \ac{ss} after hazard detection. 

        The structure of this article is as follows. Section \ref{sec:2} provides an introduction to the input required for the methodology, illustrated with a case example. Section \ref{sec:3} outlines the newly proposed methodology, ensuring \ac{fusa} compliance during the testing and actual performance. Section \ref{sec:4} details the implementation of this methodology on a real \ac{av}. The article concludes in Section \ref{sec:5} with key learnings and directions for future work.
    
\section{Problem Statement} \label{sec:2}

    This section outlines the requirements for the conversion methodology and is detailed with an example for a Renault Mégane. For the purposes of this article, these inputs are assumed to be provided by the carmaker. 

    \subsection{\ac{odd}, \ac{hara} \& \ac{tsc}}

        The initial step in \ac{fusa} analysis involves describing the vehicle's operational conditions, encapsulated by the concept of \acf{odd}. This includes elements such as speed limits, anticipated weather conditions, \acp{os}, and potential interactions with various obstacles. It sets the foundation for the definition of the system requirements for the \ac{fusa} standard, ISO~26262.
        
        Subsequently, the \ac{hara} is performed. In this study, the risk assessment of one of the items of the \ac{av}, I\_01, is summarized in Table~\ref{tab:HARA}. There are two hazards, HZ\_01 and HZ\_02, that can cause an unreasonable risk due to the item malfunction. They have been carefully studied in the \ac{hara}. Both hazards are analyzed for three different \acp{os} introduced in the \ac{odd}: $\text{OS}_1$, $\text{OS}_2$ and $\text{OS}_3$. Exposure, controllability, and severity are calculated for each combination of HZ\_ID and $\text{OS}_{\text{ID}}$, generating an ASIL. \Acp{sg} indicate what must be achieved to guarantee \ac{fusa}. They are qualitative and focus on preventing unreasonable risk due to hazards caused by malfunctioning behavior of electrical and electronic systems. The manufacturer has defined a different \ac{sg} for each hazard, SG\_01 and SG\_02. Finally, \acp{ss} are a mode of operation that a system enters or maintains to achieve a \ac{sg}. These \acp{ss} are defined in the \ac{tsc} phase of the ISO~26262, and need a technical understanding of the \ac{av} for it to reach a specific \ac{sg}. Therefore, \acp{ss} require custom design for the various \acp{os}, culminating in actions that meet \ac{fusa} requirements. 

        \begin{table}[ht]
            \small
            \centering
            \caption{Hazard analysis and risk assessment adding safety state information from the technical safety concept.}
            \label{tab:HARA}
            \resizebox{0.75\columnwidth}{!}{\begin{tabular}{|K{0.7cm}|K{1.1cm}|K{1.5cm}|K{0.75cm}|K{0.9cm}|K{0.9cm}|}
                \hline
                \rowcolor{renault!30} \multicolumn{6}{|c|}{\textbf{\ac{av} \ac{hara} + \ac{ss}}} \\
                \hline
                \hline
                \rowcolor{renault!10} {\textbf{Item ID}} & {\textbf{Hazard ID}} & {\textbf{Operating Scenario ID}} & {\textbf{ASIL}} & {\textbf{Safety Goal ID}} & {\textbf{Safety State ID}} \\
                \hline
                \multirow{6}{*}{I\_01} & \multirow{3}{*}{HZ\_01} & $\text{OS}_1$ & C & SG\_01 & SS\_01 \\
                & & $\text{OS}_2$ & C & SG\_01 & SS\_01 \\
                & & $\text{OS}_3$ & C & SG\_01 & SS\_01 \\
                \cline{2-6}
                & \multirow{3}{*}{HZ\_02} & $\text{OS}_1$ &  B & SG\_02 & SS\_02 \\
                & & $\text{OS}_2$ & C & SG\_02 & SS\_03 \\
                & & $\text{OS}_3$ & D & SG\_02 & SS\_04 \\
                \hline
            \end{tabular}}
        \end{table}
        
        The case study presented in this research is derived from a specific segment of the comprehensive \ac{hara} to facilitate a clearer understanding of the methodology. The focus is on the item I\_01, which pertains to the \ac{av} control \ac{sw} module responsible for lateral and longitudinal control. HZ\_02, associated with this item, indicates \textit{`failure of the \ac{av} control to adhere to the navigation specifications, including road speed limits and navigation reference trajectories'}.

        A significant \ac{os}, labeled $\text{OS}_3$, represents the most critical scenario and is assigned with the highest \ac{asil}. This scenario occurs when the vehicle operates on a one-lane road in the middle of high traffic, with pedestrians trying to cross, and adverse weather conditions. The SG\_02, aims to \textit{`Ensure that autonomous driving control maintains lane integrity accurately and adjusts vehicle speed to prevent collisions under all operating conditions'}. Possible \acp{ss} to reach SG\_02 include initiating an immediate \ac{av} safety stop or engaging in a degraded mode where the \ac{av} continues to drive autonomously with slight adjustments to mitigate the identified risk.

    \subsection{\acl{fta}} 

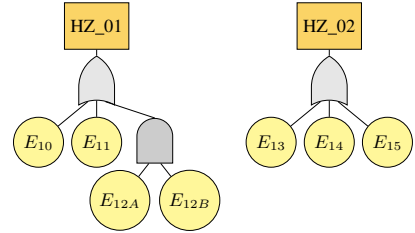
\begin{figure}[ht]
            \small
            \centering
            \resizebox{0.6\columnwidth}{!}{\begin{tikzpicture}[circuit logic US]               
                \node[rectangle, draw, fill=renault!60, minimum size=8mm] (hz2_a) at (4.5,-3) {HZ\_01};
                \node[or gate, logic gate inputs=nnn, draw, point up, fill=black!10] (or5) at (4.5,-4) {};
                \node[rectangle, draw, fill=renault!60, minimum size=8mm] (hz2_b) at (8.5,-3) {HZ\_02};
                \node[or gate, logic gate inputs=nnn, draw, point up, fill=black!10] (or6) at (8.5,-4) {};
                \node[and gate, logic gate inputs=nn, draw, point up, fill=black!20] (and1) at (5.5,-5) {};
    
                \node[circle, draw, fill=yellow!50, minimum size=8mm] (event10) at (3.5,-5) {$E_{10}$};
                \node[circle, draw, fill=yellow!50, minimum size=8mm] (event11) at (4.5,-5) {$E_{11}$};
                \node[circle, draw, fill=yellow!50, minimum size=8mm] (event12a) at (4.9,-6) {$E_{12A}$};
                \node[circle, draw, fill=yellow!50, minimum size=8mm] (event12b) at (6.1,-6) {$E_{12B}$};
                \node[circle, draw, fill=yellow!50, minimum size=8mm] (event13) at (7.5,-5) {$E_{13}$};
                \node[circle, draw, fill=yellow!50, minimum size=8mm] (event14) at (8.5,-5) {$E_{14}$};
                \node[circle, draw, fill=yellow!50, minimum size=8mm] (event15) at (9.5,-5) {$E_{15}$};
    
                \draw (or5.output) -- (hz2_a);
                \draw (or6.output) -- (hz2_b);
                \draw (event10) -- (or5.input 1);
                \draw (event11) -- (or5.input 2);
                \draw (and1.output) -- (or5.input 3);
                \draw (event13) -- (or6.input 1);
                \draw (event14) -- (or6.input 2);
                \draw (event15) -- (or6.input 3);
                \draw (event12a) -- (and1.input 1);
                \draw (event12b) -- (and1.input 2);
            \end{tikzpicture}}
            \caption{Fault Tree Analysis for hazards related to I\_01: HZ\_01 and HZ\_02.}
            \label{fig:FTA}
        \end{figure}

        \Ac{fta} \cite{vesely1981fault} is a top-down technique that, in the context of ISO~26262, has been commonly used to deduce possible failures of vehicle systems from potential hazards (\ac{hara}), system description and \acp{sg}. \Ac{fta} is also useful for other purposes, such as deriving \acp{fsr} from \acp{sg} \cite{schonemann2019fault} or calculating the probability of the occurrence of a failure event. An example of a fault tree, obtained from the \ac{hara} of Table \ref{tab:HARA}, is shown in Fig.~\ref{fig:FTA}. The logical gates in the tree illustrate how different failures combine to generate hazards. 

        \Ac{fta} helps identify what could fail, but does not specify what safety measures or performance levels are necessary. Integrating \ac{hara} with \ac{fta} provides a detailed analysis of potential failures and their link to hazards, fundamental for the novel supervision architecture based on \acp{bt}.

        For the FTA example of Fig.~\ref{fig:FTA}, the correspondent probabilities of event occurrence are presented in Table~\ref{table:probabilities_fta}. They are obtained from experimental results under nominal conditions.

        \begin{table}[ht]
            \centering
            \small
            \renewcommand{\arraystretch}{1.35} 
            \caption{Event identification probabilities}
            \resizebox{0.95\columnwidth}{!}{\begin{tabular}{|c|c|c|c|c|c|c|c|}
                \hline
                \multicolumn{8}{|c|}{\cellcolor{renault!20}\textbf{$\mathbf{E_{i}}$ Probability of Occurrence}} \\ 
                \hline
                \cellcolor{renault!10}\textbf{i} & \cellcolor{renault!10}10 & \cellcolor{renault!10}11 & \cellcolor{renault!10}12A & \cellcolor{renault!10}12B & \cellcolor{renault!10}13 & \cellcolor{renault!10}14 & \cellcolor{renault!10}15\\ 
                \hline
                \cellcolor{renault!10}\textbf{$\mathbf{p}$} & $5\cdot 10^{-4}$ & $1\cdot 10^{-3}$ & $1\cdot 10^{-4}$ & $2\cdot 10^{-4}$ & $2\cdot 10^{-3}$ & $5\cdot 10^{-4}$ & $1\cdot 10^{-3}$ \\ 
                \hline
            \end{tabular}}           
            \label{table:probabilities_fta}
        \end{table} 

        Integrating \ac{hara} with \ac{fta} provides a detailed analysis of potential failures and their link to hazards. Defining \acp{sg} and \acp{ss}, ensuring system-level safety considerations, and engaging in continuous safety management are essential to fully comply with the functional safety requirements of ISO~26262.

\section{Method} \label{sec:3}

    Semi-Markov models have been used to convert static fault trees \cite{vesely1981fault}, generated from \ac{fta}, into dynamic trees \cite{dugan1992dynamic}, \cite{volk2018fast}, \cite{kabir2019conceptual}. This conversion and the similarity of the hierarchy between \acp{bt} and fault trees indicate that \acp{bt} can also transform static fault trees into dynamic ones. In this case, the benefits of this novel approach include intuitive visualization and adaptability.

\begin{figure}[ht]
        \centering
        \includegraphics[width=0.48\textwidth]{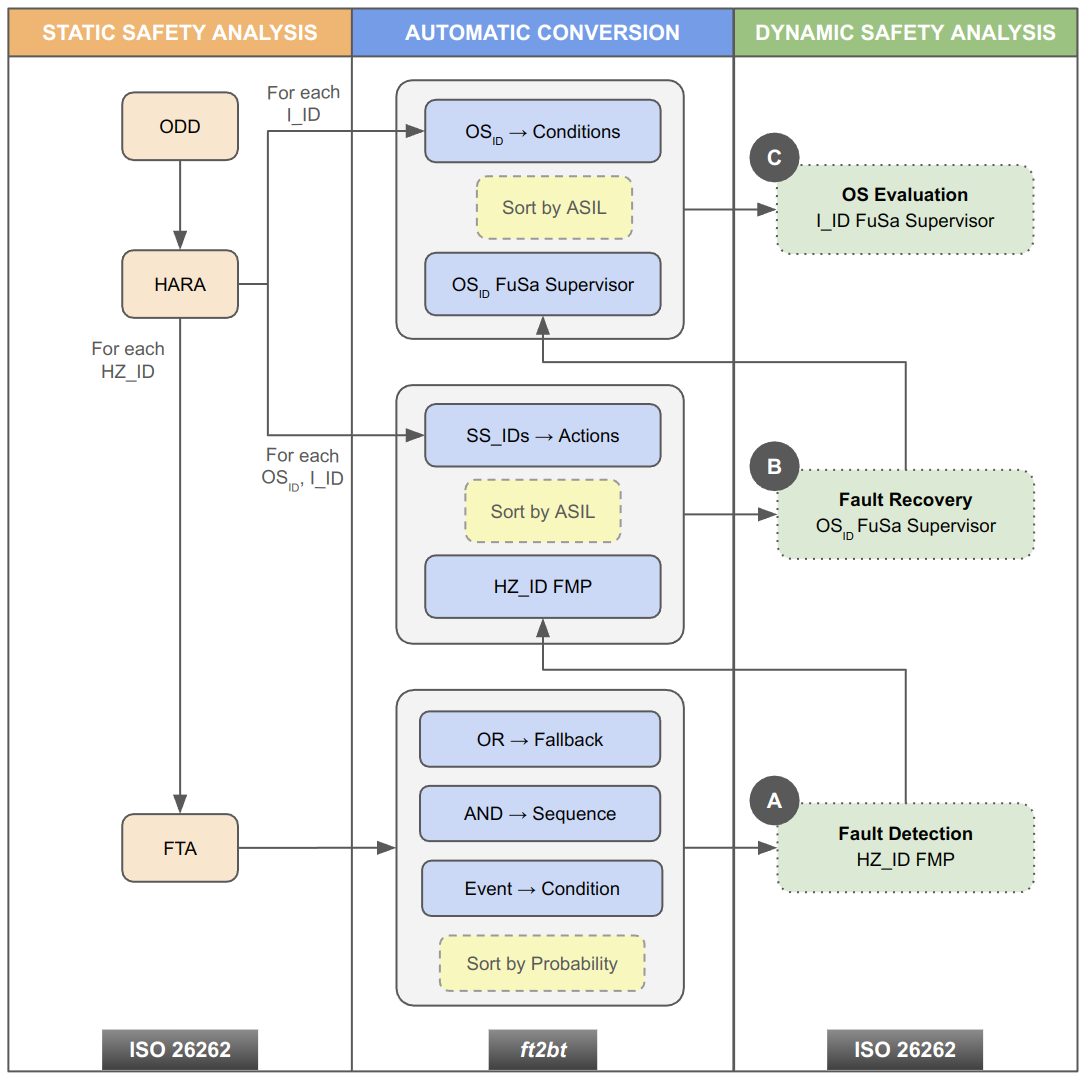}
        \caption{Block diagram of the translation from static into dynamic safety analysis proposal. The first phase, indicated with (A) consists of converting fault trees into \acp{fmp}. Subsequently, the second phase of the translation, (B), gets the information from the \ac{hara} and from the already designed \acp{fmp} to generate the \ac{fusa} supervisor for each \ac{os} defined in the \ac{odd}. Finally, the last step, (C), consists of finding the current \ac{os}, and linking it with the correspondent \ac{os} \ac{fusa} supervisor.}
        \label{fig:system_figure}
    \end{figure}

    Therefore, a novel supervision methodology for \acp{av}, based in \acp{bt}, is presented in this section. It automatically converts a static safety analysis into a dynamic run-time assessment. The result is a group of supervisors for \acp{av}, specifically adapted for each item, I\_ID, included in the \ac{hara}. They are in compliance with \ac{fusa} requirements, as they are based on the \ac{hara} and \ac{fta}. The process is represented in the block diagram of Fig.~\ref{fig:system_figure}. \Ac{bt} elements are described in Table~\ref{table:behavior_tree}.

    \begin{table}[ht]
        \centering
        \small
        \caption{Symbols and description of the elements of a behavior tree}
        \label{table:behavior_tree}
        \resizebox{0.8\columnwidth}{!}{\begin{tabular}{|K{1.6cm}|K{1.2cm}|K{4.6cm}|}
            \hline
            \textbf{Node Type} & \textbf{Symbol} & \textbf{Description} \\
            \hline
            Fallback & \centering\tikz{\node[fallback] {?};} & Activates the first child node that succeeds, often used to find the first action that can be taken. \\
            \hline
            Sequence & \centering\tikz{\node[sequence] {$\rightarrow$};} & Executes its child nodes in order, moving to the next only when the current one succeeds. \\
            \hline
            Parallel & \centering\tikz{\node[parallel] {$\rightrightarrows$};} & Executes all of its children in parallel and succeeds if a certain number of them succeed. \\
            \hline
            Decorator & \centering\tikz{\node[decorator] {$\delta$};} & Modifies the behavior of its child node in various ways, such as inverting success/failure. \\
            \hline
            Action & \centering\tikz{\node[state] {ID};} & Performs a specific task or action, usually represents the leaf nodes of the tree. \\
            \hline
            Condition & \centering\tikz{\node[condition] {ID};} & Checks a specific condition and succeeds if the condition is met. \\
            \hline
            Subtree & \centering\tikz{\node[subtree, fill=white] {ID};} & Encapsulates a segment that represents a complex behavior as a single node. \\
            \hline
        \end{tabular}}
    \end{table}

    The first step of the method (Subsection~\ref{sec:A}) consists in translating the fault trees into \acp{fmp} for run-time fault detection. The second step of the technique (Subsection~\ref{sec:B}) considers \ac{hara} information to guarantee \ac{fusa}, given an \ac{os} and an identified risk. Thus, the connection between hazards and their correspondent \acp{ss} is encompassed in this phase of the supervisor design. Finally, the last step of the conversion (Subsection~\ref{sec:C}) is in charge of finding the \ac{av}'s \ac{os} to select its correspondent architecture to ensure safety. It extracts information from the \ac{hara}. All the mentioned phases are based on the \ac{bt} architecture.

    \subsection{Behavior Trees for Fault Detection (\acp{fmp})} \label{sec:A}
        \subsubsection{\ac{fta} elements into \ac{bt} nodes conversion}

            The first step in transforming a static safety analysis into a dynamic one involves converting a fault tree, developed during the \ac{fta} phase, into a behavior tree. The hierarchical structures of both and the similarities in their elements facilitate an intuitive transition from static to dynamic representation. This section details how logic gates from fault trees can be effectively modeled as nodes in \acp{bt}, while preserving the original hierarchy. This process ensures that the logical relationships and dependencies established in \ac{fta} are accurately reflected and maintained in the dynamic \ac{bt} framework.

            Both the \textit{AND gate} in \ac{fta} and the \textit{sequence node} in a \ac{bt} operate on the principle that all components or conditions must be satisfied for success. However, they differ in functionality. The \textit{AND gate} statically evaluates all inputs simultaneously, requiring them to be true for a true output. In contrast, a \ac{bt} \textit{sequence node} executes the child nodes in order, succeeding only if each child node does, allowing for adaptability to changing conditions due to its dynamic and ordered nature.

            The \textit{OR gate} in \ac{fta} and the \textit{fallback node} in a \ac{bt} also share conceptual similarity, but differ in execution. The \textit{OR gate} outputs true if any of its inputs is true, evaluating all inputs simultaneously and regardless of order. In contrast, the \textit{fallback node} in a \ac{bt} executes its child nodes sequentially, proceeding to the next child only if the current node fails and succeeds if any child node succeeds.
    
            In \ac{fta}, events are specific occurrences or states that determine the probability of system failure, represented as binary static nodes that focus on the causality of the failure. Contrarily, \ac{bt} \textit{condition nodes} are dynamic states or prerequisites that guide the execution of subsequent tasks. Although both are key to their structures and influence the outcomes, \ac{fta} events are static and fail-focused, whereas \ac{bt} \textit{condition nodes} dynamically direct behavioral logic in real-time systems.
    
\begin{figure}[ht]
                \small
                \centering
                \resizebox{0.75\columnwidth}{!}{\begin{tikzpicture}[
                level 1/.style={sibling distance=10mm, level distance=10mm},
                level 2/.style={sibling distance=12.5mm, level distance=12.5mm},
                level 3/.style={sibling distance=12.5mm, level distance=10mm},
                level 4/.style={sibling distance=12.5mm, level distance=12.5mm},]        
                
                \begin{scope}[xshift=2.25cm] 
                    \node[action] (b) {HZ\_02}
                        child{node[fallback] {?}
                            child {node[condition] {$E_{13}$}}
                            child {node[condition] {$E_{15}$}}
                            child {node[condition] {$E_{14}$}}
                        };
                \end{scope}
                
                \begin{scope}[xshift=-2.25cm] 
                    \node[action] (a) {HZ\_01}
                        child{node[fallback] {?}
                            child {node[condition] {$E_{11}$}}
                            child {node[condition] {$E_{10}$}}
                            child{node[sequence] {$\rightarrow$}
                                child {node[condition] {$E_{12A}$}}
                                child {node[condition] {$E_{12B}$}}
                            }
                        };
                \end{scope}
                \end{tikzpicture}}
                \caption{Subtrees of hazards HZ\_01 and HZ\_02 fault detection, extracted from the translation of the \ac{fta} presented in Fig.~\ref{fig:FTA}. They represent the \acp{fmp} for both hazards. The probabilities of Table~\ref{table:probabilities_fta} have been considered for the order of event identification.}
                \label{fig:BT_hazard}
            \end{figure}
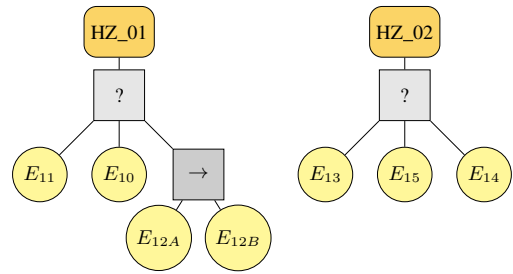

            Due to the explained similarity of both static and dynamic safety analyses, the conversion between \ac{fta} \textit{AND}/\textit{OR} logic gates into \ac{bt} \textit{sequence}/\textit{fallback nodes} and between events and \textit{condition nodes} is direct and, therefore, is automatized. An example of the \ac{bt} for HZ\_01 and HZ\_02 detection is represented in Fig.~\ref{fig:BT_hazard} as a translation of the previously introduced \ac{fta} of Fig.~\ref{fig:FTA}.

        \subsubsection{Prioritization of Event Identification by probability of occurrence}
            
            Given the sequential nature of \acp{bt}, it is advisable to carefully consider the order of the \textit{condition nodes} represented. It decreases the detection latency for the most probable causes, and reduces resource consumption.
            
            The order of events that can cause a hazard should be based on the probabilities derived from \ac{fta}. Therefore, within a \ac{bt}, the most probable event associated with a particular hazard should be placed on the left side of a \textit{fallback node} and on the right side of a \textit{sequence node}, as represented in Fig.~\ref{fig:BT_hazard}. This distribution ensures that higher-risk events are prioritized and evaluated promptly in the case of a \textit{fallback node}, where only one \texttt{SUCCESS} results in a hazard identification. Alternatively, in the case of a \textit{sequence node} it is advisable to first check the events with lower probability, as only one \texttt{FAILURE} is sufficient to ensure that the hazard is not caused by this part of the \ac{bt}.

    \subsection{Behavior Trees for Fault Recovery (\ac{os} \ac{fusa} Supervisors)}\label{sec:B}
        \subsubsection{\ac{hara} information into \ac{bt} nodes conversion}
        
            The second phase of the conversion from static to dynamic modeling aims to generate a fault recovery architecture. Based on the \ac{hara} information, it relates hazards with \acp{ss} for each \ac{os} of the same item. 
            
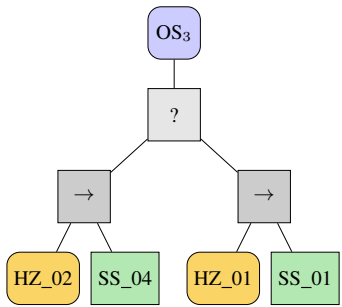
\begin{figure}[ht]
                \small
                \centering
                \resizebox{0.5\columnwidth}{!}{\begin{tikzpicture}[
                level 1/.style={sibling distance=35mm, level distance=12.5mm},
                level 2/.style={sibling distance=27.5mm, level distance=12.5mm},
                level 3/.style={sibling distance=12.5mm, level distance=12.5mm}
            ]
                    \node[subtree] {$\text{OS}_3$}
                        child {node[fallback] {?}
                            child {node[sequence] {$\rightarrow$}
                                child {node[action] {HZ\_02}}
                                child{node[state] {SS\_04}}
                            }
                            child {node[sequence] {$\rightarrow$}
                                child {node[action] {HZ\_01}}
                                child{node[state] {SS\_01}}
                            }
                        };
                \end{tikzpicture}}
                \caption{Structure of the fault recovery \ac{bt}, specifically for the scenario $\text{OS}_3$ of item I\_01. It is directly extracted from the \ac{hara} information, described in Table~\ref{tab:HARA}. Hazard detection subtrees, HZ\_01 and HZ\_02, are detailed in Fig.~\ref{fig:BT_hazard}. Hazards are prioritized by \ac{asil}.}
                \label{fig:BT_recovery} 
            \end{figure}

            The design of the \ac{bt} is detailed from a top-down perspective. In the first layer, a \textit{fallback node} assesses the occurrence of hazards, HZ\_01 or HZ\_02. In the absence of such events, no action is executed, signifying that the \ac{av} behavior remains unchanged as per \ac{fusa} requirements. The second layer incorporates a \textit{sequence node} for each hazard analyzed. These nodes are tasked with executing the appropriate \ac{ss} in response to identified hazards. Fig.~\ref{fig:BT_recovery} represents the fault recovery structure of the hazards, detected with the \acp{fmp}, related to item I\_01 and $\text{OS}_3$.

        \subsubsection{Prioritization of Hazard \acp{fmp} by \ac{asil}}
            In this phase of the automatic conversion, it is also important to consider the sequential performance of \acp{bt}. Consequently, it is necessary to identify the most critical hazards and place them on the left side of the \ac{bt}. 
            
            The priority of a hazard is determined by extracting the \ac{asil} associated with the \ac{os} and item in the \ac{hara}. Thus, the initial hazard to be monitored represents the greatest risk to the \ac{av} (highest \ac{asil}). In instances where multiple hazards share the same maximum \ac{asil}, the ordering should be based on the \ac{fta} probabilities associated with each hazard, with the most likely hazard being prioritized in the sequence. In accordance with the assessment from Table~\ref{tab:HARA}, HZ\_02 is identified as the most critical hazard for $\text{OS}_3$ and item I\_01, particularly because it involves an ASIL~D rating. This prioritization, shown in Fig.~\ref{fig:BT_recovery}, ensures that the most safety-critical hazards are addressed promptly within the \ac{bt} structure.

    \subsection{Behavior Trees for Operating Scenario Evaluation} \label{sec:C}
        \subsubsection{\ac{hara} information into \ac{bt} nodes conversion}
            The result of the implementation and integration of the two previous phases of the methodology generates a \ac{fusa} supervisor \ac{bt} capable of dynamically evaluating the \ac{av} safety of a certain item and a specific \ac{os}. 
    
\begin{figure}[ht]
                \small
                \centering
                \resizebox{0.7\columnwidth}{!}{\begin{tikzpicture}[
                level 1/.style={sibling distance=35mm, level distance=12.5mm},
                level 2/.style={sibling distance=25mm, level distance=12.5mm},
                level 3/.style={sibling distance=12.5mm, level distance=12.5mm}
            ]
                    \node[root, node distance=2.5cm] {I\_01}
                        child{node[fallback] {?}
                            child {node[sequence] {$\rightarrow$}
                                child {node[condition] {$\text{OS}_3$}}
                                child{node[subtree] {$\text{OS}_3$}}
                            }
                            child {node[sequence] {$\rightarrow$}
                                child {node[condition] {$\text{OS}_2$}}
                                child{node[subtree] {$\text{OS}_2$}}
                            }
                            child {node[sequence] {$\rightarrow$}
                                child {node[condition] {$\text{OS}_1$}}
                                child{node[subtree] {$\text{OS}_1$}}
                            }
                        };
                \end{tikzpicture}}
                \caption{\ac{bt} architecture for the \ac{os} evaluation of item I\_01, based on the \acp{os} contemplated in the \ac{odd} and \ac{hara} of Table~\ref{tab:HARA}. First, the \ac{bt} checks the \ac{os} and subsequently performs the \ac{fmp} and fault recovery processes, as described in Fig.~\ref{fig:BT_recovery}. \Acp{os} are prioritized by \ac{asil}.}
                \label{fig:BT_fusa_supervisor}
            \end{figure}
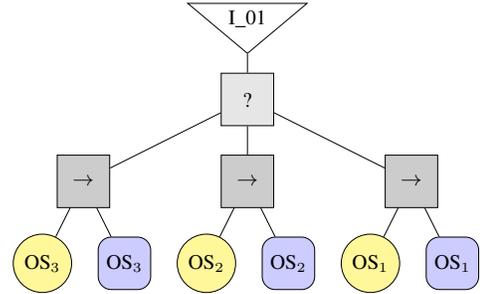

            This last step of the supervisor design aims to find the current \ac{os} to link it to the correspondent fault recovery \ac{bt}. In this structure, the \textit{condition nodes} evaluate the \ac{os}, the \textit{subtree nodes} have previously been designed as in Fig.~\ref{fig:BT_recovery}. Thus, when the scenario is identified, the fault recovery \ac{bt} associated to the \ac{os} and item is executed, thanks to the \textit{sequence nodes} of the \ac{bt}. The \ac{os} evaluation \ac{bt}, based in the \ac{hara} information, is represented in Fig.~\ref{fig:BT_fusa_supervisor}.
    
        \subsubsection{Prioritization of \ac{os} identification by \ac{asil}}

            In the last phase of the methodology, which corresponds to the high-level architecture of the \ac{bt}, the order of the \ac{os} \textit{condition nodes} is important to optimize the \ac{fusa} supervision. The \ac{os} that contains the highest \ac{asil} in the \ac{hara} is prioritized, being placed on the left side of the tree.

    \subsection{Dynamic Safety Analysis}

\begin{figure}[ht]
            \centering
            \includegraphics[width=0.45\textwidth]{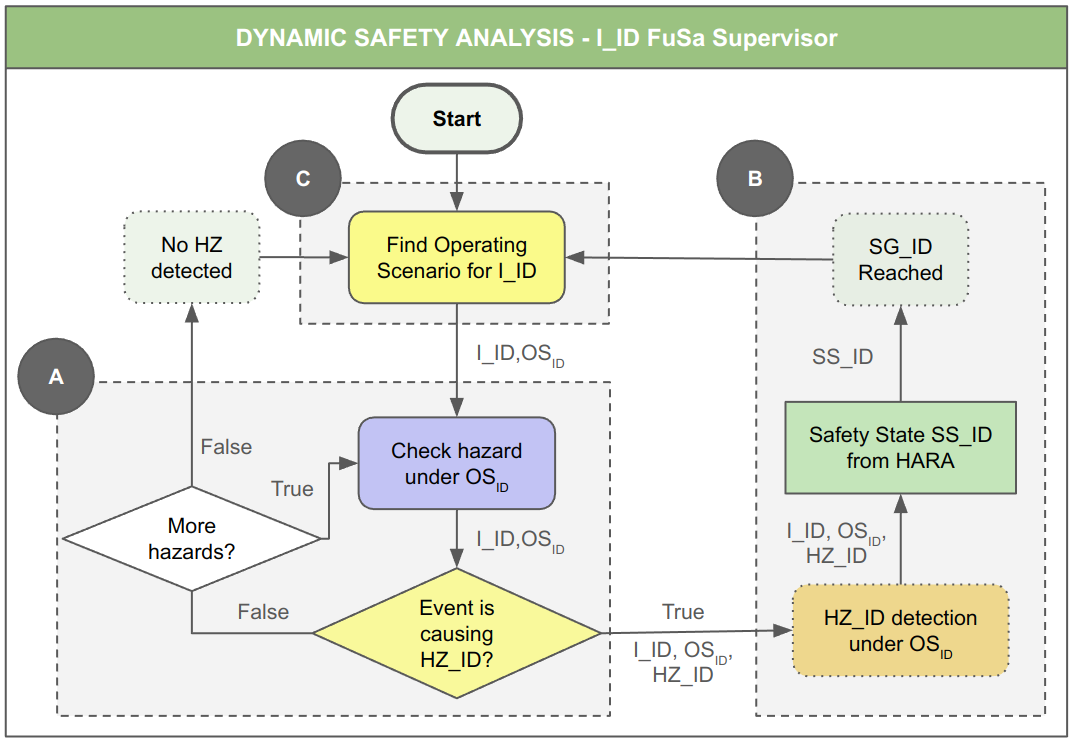}
            \caption{Flowchart of the \ac{fusa} supervisor behavior for each item in the \ac{hara}. (A), (B) and (C) belong to blocks of the dynamic safety analysis in Fig.~\ref{fig:system_figure}.}
            \label{fig:flowchart}
        \end{figure}

        Once the \ac{fusa} supervisor for each item in the \ac{hara} is designed, the run-time process can be described. The underlying logic of the dynamic safety analysis is illustrated in Fig.~\ref{fig:flowchart}. 

        The initial section of the supervisor aims to determine the current $\text{OS}_{\text{ID}}$ for a specific \ac{av} item, I\_ID, as elaborated in Section~\ref{sec:C}. Subsequently, after the identification of the $\text{OS}_{\text{ID}}$, the hazards associated with both the \ac{os} and the item are evaluated using the \ac{fmp}, detailed in Section~\ref{sec:A}. If no hazards are detected, the \ac{fusa} supervisor initiates the process again, beginning with \ac{os} selection. In contrast, if a hazard-related event is identified, the fault recovery architecture takes charge, implementing the corresponding \ac{ss} until the \ac{sg} is achieved. Once the \ac{av} is considered safe, the safety action that recovers the vehicle ceases, and the supervision cycle resumes.

\section{Results} \label{sec:4}
        
    Upon presenting the information on the safety standards in Section~\ref{sec:2} and detailing the methodology to convert from static to run-time safety analysis, the next step involves its implementation in an autonomous Renault Mégane. The aim of this section is to assess the accuracy of event identification and the execution of the correspondent \ac{ss} by the \ac{bt}. Tests are performed to evaluate the detection latency and to confirm that the actions taken (\acp{ss}) ensure the safety of the \ac{av}.

    \begin{figure*}[!t]
        \centering
        \includegraphics[width=0.7\textwidth]{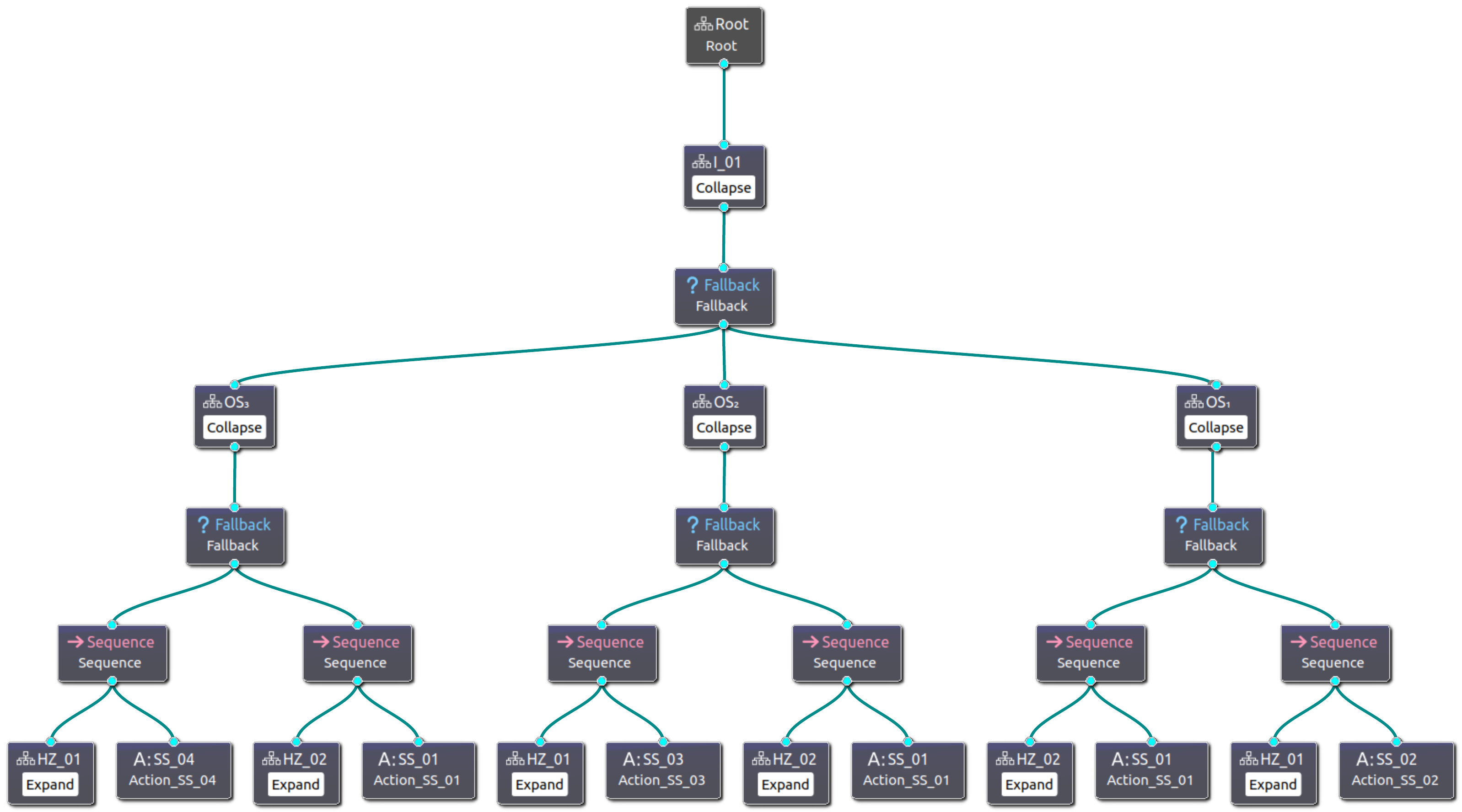}
        \caption{Integration of the \ac{fusa} supervisor with \acp{bt}, using the Groot \ac{gui}. The hazard subtrees, HZ\_01 and HZ\_02, are replicated on each \ac{os}. The \acp{bt} are configured to automatically recognize that subtrees sharing the same identification correspond to the same event architecture.}
        \label{fig:groot}
    \end{figure*}

    \subsection{Real Implementation}
        To facilitate the integration of the \ac{fusa} supervisor \acp{bt} within the existing \ac{sw} of the \ac{av}, the \ac{ros} framework and Groot \ac{gui} are used together with the BehaviorTree.CPP library. The software tool designed to automatize the translation from \ac{fta}/\ac{hara} to \acp{bt}, \textit{ft2bt}\footnote{\href{https://github.com/cconejob/ft2bt_converter}{https://github.com/cconejob/ft2bt\_converter}}, is also essential to automatically generate the supervisor's architecture.

        The input required for the automated conversion process is the XML file of the \ac{fta}, extracted from a specific diagramming tool\footnote{\href{https://app.diagrams.net/?src=about}{https://app.diagrams.net/?src=about}}, and the CSV file of the \ac{hara}. The prioritization of hazards, events, and actions follows the guidelines established in Section~\ref{sec:3}. Once the \ac{bt} is generated, as represented in Fig.~\ref{fig:groot}, it must be imported into the \ac{av} \ac{sw} using the ROS BehaviorTree.CPP library. 

        Subsequently, each \ac{bt} \textit{condition node} (Event) must be meticulously designed in the \ac{sw} to ensure its precise identification. This requires a comprehensive understanding of both the \ac{hw} and \ac{sw} development processes of the \ac{av} to accurately define the rules necessary for the detection of events (fault detection or \ac{os} identification). Similarly, actions need to be specifically designed to ensure the effectiveness of the \ac{ss} until the \ac{sg} is achieved. In contrast, the functionalities of the \textit{fallback} and \textit{sequence nodes} are inherently defined in the \ac{bt} and, thus, do not require explicit specification.

    \subsection{Testing}
        After the design and integration of \ac{fusa} supervisor \acp{bt} in the actual \ac{sw} pipeline of the Renault Mégane, the subsequent phase involves testing the \ac{av} under nominal conditions. This testing phase is important due to two interconnected objectives: evaluating the performance of the \ac{fmp} during nominal tests and refining the parameters for event identification. 

        The primary objective involves evaluating the autonomous mode of the Renault Mégane with the supervisory system activated. During these evaluations, the emergence of several false positives was observed, highlighting the need for parameter adjustment. For example, the event $E_{13}$, which denotes the condition \textit{'current speed exceeds the road limit or the navigation reference'}, initially generated many false positives. The \ac{av}'s speed persistently surpassed the navigation reference immediately following the \ac{sw}'s recognition that a reduction in speed was necessary. This scenario can appear in the presence of an obstacle or a sharp curve. For this reason, the parameter adjustment must be performed for $E_{13}$. A low value of the admissible anomaly time, $t_\text{anomaly}$, would induce unnecessary false positives, while a high value would detect failures too late for safety purposes. Moreover, a speed tolerance should also be set if necessary.

        Evaluations are carried out through a combination of simulation and practical tests. Initially, all safety algorithms are tested within a simulation environment. This was followed by practical tests conducted with the Renault Mégane \ac{av} at the Renault facilities in Spain for different \acp{os}.

    \subsection{Falsification}
        Falsification targets specific scenarios where the model fails to meet the \ac{fusa} criteria. This process uncovers the response of the methodology to safety-critical faults within the system, helping to improve its overall efficacy and delineate its limitations.

        These tests involve operating under nominal conditions before deliberately inducing specific faults, each corresponding to a distinct event. Initially, isolated failures (47 tests) are introduced, followed by simultaneous testing of multiple failures (33 tests). Similarly to the approach in the testing phase, simulation-based falsification precedes real falsification experiments with the Renault Mégane.

        The results of the falsification tests for various events on the real platform are presented in the confusion matrix in Table \ref{table:confusion_matrix_real}. This matrix illustrates the percentage of instances where the supervisor's event identification (predicted values) aligns with or diverges from the ground truth (actual values). 

        \begin{table}[ht]
            \centering
            \small
            \renewcommand{\arraystretch}{1.35} 
            \caption{Confusion matrix for the field monitoring process testing \\ and falsification in the autonomous Renault Mégane}
            \resizebox{0.7\columnwidth}{!}{\begin{tabular}{|cc|c|c|c|c|c|c|c|c|}
                \cline{3-10}
                \multicolumn{2}{c|}{} & \multicolumn{8}{c|}{\cellcolor{renault!20}\textbf{Predicted Event}} \\ 
                \cline{3-10}
                \multicolumn{2}{c|}{} &  \cellcolor{renault!10}Nom. & \cellcolor{renault!10}10 & \cellcolor{renault!10}11 & \cellcolor{renault!10}12A & \cellcolor{renault!10}12B & \cellcolor{renault!10}13 & \cellcolor{renault!10}14 & \cellcolor{renault!10}15 \\ 
                \hline
                \multicolumn{1}{|c|}{\cellcolor{blue!20}} & \cellcolor{blue!10}Nom. & \cellcolor{green!10} 98\% &  \cellcolor{red!20} 0\% &  \cellcolor{red!20} 0\% &  \cellcolor{red!20} 0\% &  \cellcolor{red!20} 0\% &  \cellcolor{red!20} 0\% &  \cellcolor{red!20} 2\% & \cellcolor{red!20} 0\% \\ 
                \cline{2-10}
                \multicolumn{1}{|c|}{\cellcolor{blue!20}} & \cellcolor{blue!10}10 &  \cellcolor{red!30} 0\% & \cellcolor{green!10} 100\% &  \cellcolor{red!20} 0\% &  \cellcolor{red!20} 0\% &  \cellcolor{red!20} 0\% &  \cellcolor{red!20} 0\% &  \cellcolor{red!20} 0\% & \cellcolor{red!20} 0\% \\ 
                \cline{2-10}
                \multicolumn{1}{|c|}{\cellcolor{blue!20}} & \cellcolor{blue!10}11 &  \cellcolor{red!30} 0\% &  \cellcolor{red!20} 0\% & \cellcolor{green!10} 100\% &  \cellcolor{red!20} 0\% &  \cellcolor{red!20} 0\% &  \cellcolor{red!20} 0\% &  \cellcolor{red!20} 0\% & \cellcolor{red!20} 0\%\\ 
                \cline{2-10}
                \multicolumn{1}{|c|}{\cellcolor{blue!20}} & \cellcolor{blue!10}12A & \cellcolor{red!30} 6\% &  \cellcolor{red!20} 0\% &  \cellcolor{red!20} 4\%  & \cellcolor{green!10} 90\% &  \cellcolor{red!20} 0\% &  \cellcolor{red!20} 0\% &  \cellcolor{red!20} 0\% &  \cellcolor{red!20} 0\% \\ 
                \cline{2-10}
                \multicolumn{1}{|c|}{\cellcolor{blue!20}} & \cellcolor{blue!10}12B &  \cellcolor{red!30} 0\% &  \cellcolor{red!20} 0\% &  \cellcolor{red!20} 0\% &  \cellcolor{red!20} 5\% & \cellcolor{green!10} 95\% &  \cellcolor{red!20} 0\% &  \cellcolor{red!20} 0\% &  \cellcolor{red!20} 0\% \\ 
                \cline{2-10}
                \multicolumn{1}{|c|}{\cellcolor{blue!20}} & \cellcolor{blue!10}13 &  \cellcolor{red!30} 0\% &  \cellcolor{red!20} 0\% &  \cellcolor{red!20} 0\% &  \cellcolor{red!20} 0\% &  \cellcolor{red!20} 6\% & \cellcolor{green!10} 88\% &  \cellcolor{red!20} 0\% &  \cellcolor{red!20} 6\% \\ 
                \cline{2-10}
                \multicolumn{1}{|c|}{\cellcolor{blue!20}} & \cellcolor{blue!10}14 &  \cellcolor{red!30} 0\% &  \cellcolor{red!20} 0\% &  \cellcolor{red!20} 0\% &  \cellcolor{red!20} 0\% &  \cellcolor{red!20} 0\% &  \cellcolor{red!20} 14\% & \cellcolor{green!10} 86\% &  \cellcolor{red!20} 0\% \\ 
                \cline{2-10}
                \multicolumn{1}{|c|}{\parbox[t]{2mm}{\multirow{-8}{*}{\rotatebox[origin=c]{90}{\cellcolor{blue!20}\textbf{Actual Event}}}}} & \cellcolor{blue!10}15 &  \cellcolor{red!30} 0\% &  \cellcolor{red!20} 0\% &  \cellcolor{red!20} 0\% &  \cellcolor{red!20} 0\% &  \cellcolor{red!20} 0\% &  \cellcolor{red!20} 3\%  &  \cellcolor{red!20} 0\% & \cellcolor{green!10} 97\% \\ 
                \hline
            \end{tabular}}           
            \label{table:confusion_matrix_real}
        \end{table} 

        Under nominal conditions, the \ac{fmp} part of the \ac{fusa} supervisor has consistently prevented the generation of false positives, accounting for 2\% of the total scenarios. In fault conditions, the supervisor accurately identified the corresponding event in an average of 94\% of falsification tests. This accuracy directly depends on the heuristic rules defined in the \ac{bt}.

        Detection latencies that exceed 0.5~s during event transitions are classified as false negatives. The average detection latencies for each event are detailed in Table \ref{table:detection_latency}. The latency in detection is influenced by the placement of the \ac{bt} for each event within the system hierarchy. Events assigned a higher priority exhibit shorter detection latencies. Consequently, an event related to a hazard classified as \ac{asil} D is detected more quickly than an \ac{asil} B hazard.

        \begin{table}[ht]
            \centering
            \small
            \renewcommand{\arraystretch}{1.35} 
            \caption{Detection latency average for each event}
            \resizebox{0.65\columnwidth}{!}{\begin{tabular}{|c|c|c|c|c|c|c|c|}
                \hline
                \multicolumn{8}{|c|}{\cellcolor{renault!20}\textbf{$\mathbf{E_{i}}$ Detection Latency}} \\ 
                \hline
                \cellcolor{renault!10}\textbf{i} & \cellcolor{renault!10}10 & \cellcolor{renault!10}11 & \cellcolor{renault!10}12A & \cellcolor{renault!10}12B & \cellcolor{renault!10}13 & \cellcolor{renault!10}14 & \cellcolor{renault!10}15\\ 
                \hline
                \cellcolor{renault!10}\textbf{$\mathbf{\overline{t}}$ [s]} & 0.115 &  0.127 & 0.135 & 0.147 & 0.098 & 0.108 & 0.102 \\ 
                \hline
            \end{tabular}}           
            \label{table:detection_latency}
        \end{table} 

\begin{figure}[ht]
            \centering
            \includegraphics[width=0.37\textwidth]{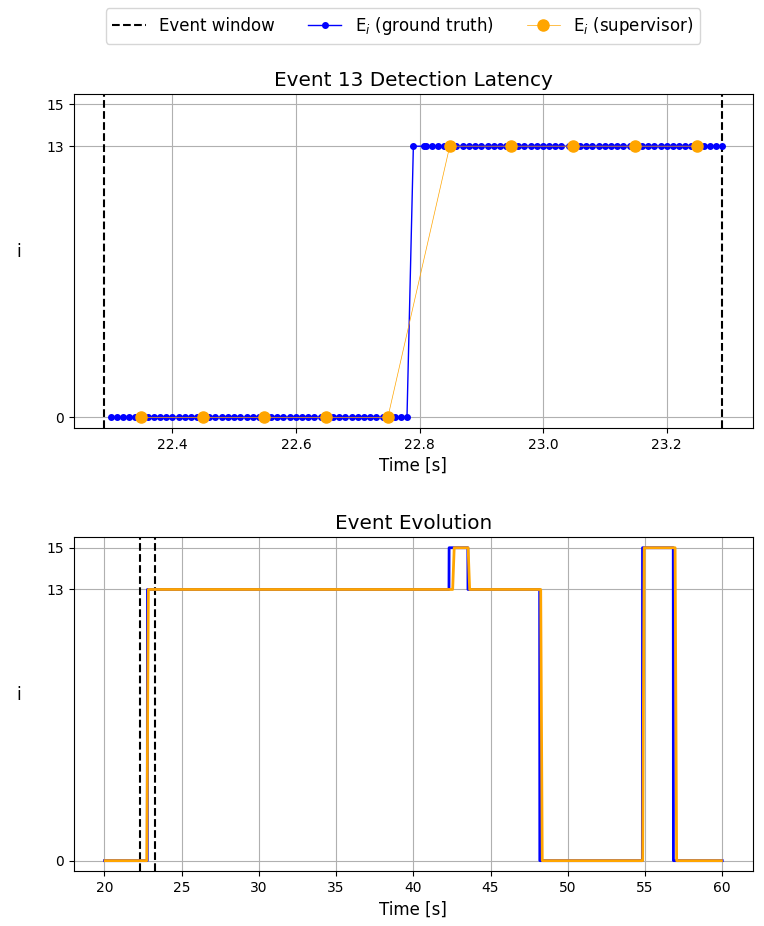}
            \caption{Experiment 63: Falsification for event $E_{13}$. The run-time event assessment is compared with the ground truth obtained from an offline analysis. The lower graph represents the evolution of the event identification during this falsification process to show the event interaction. $E_{0}$ represents the nominal conditions.}
            \label{fig:falsification_13}
        \end{figure}

        The detection latency for the falsification of $E_{13}$ is shown in Fig.~\ref{fig:falsification_13}. Ground truth data are derived from a comprehensive offline analysis of the vehicle's data (rosbags), whereas the supervisor's predictions are based on the real-time output from the \ac{fmp}.
        
\begin{figure}[ht]
            \centering
            \includegraphics[width=0.47\textwidth]{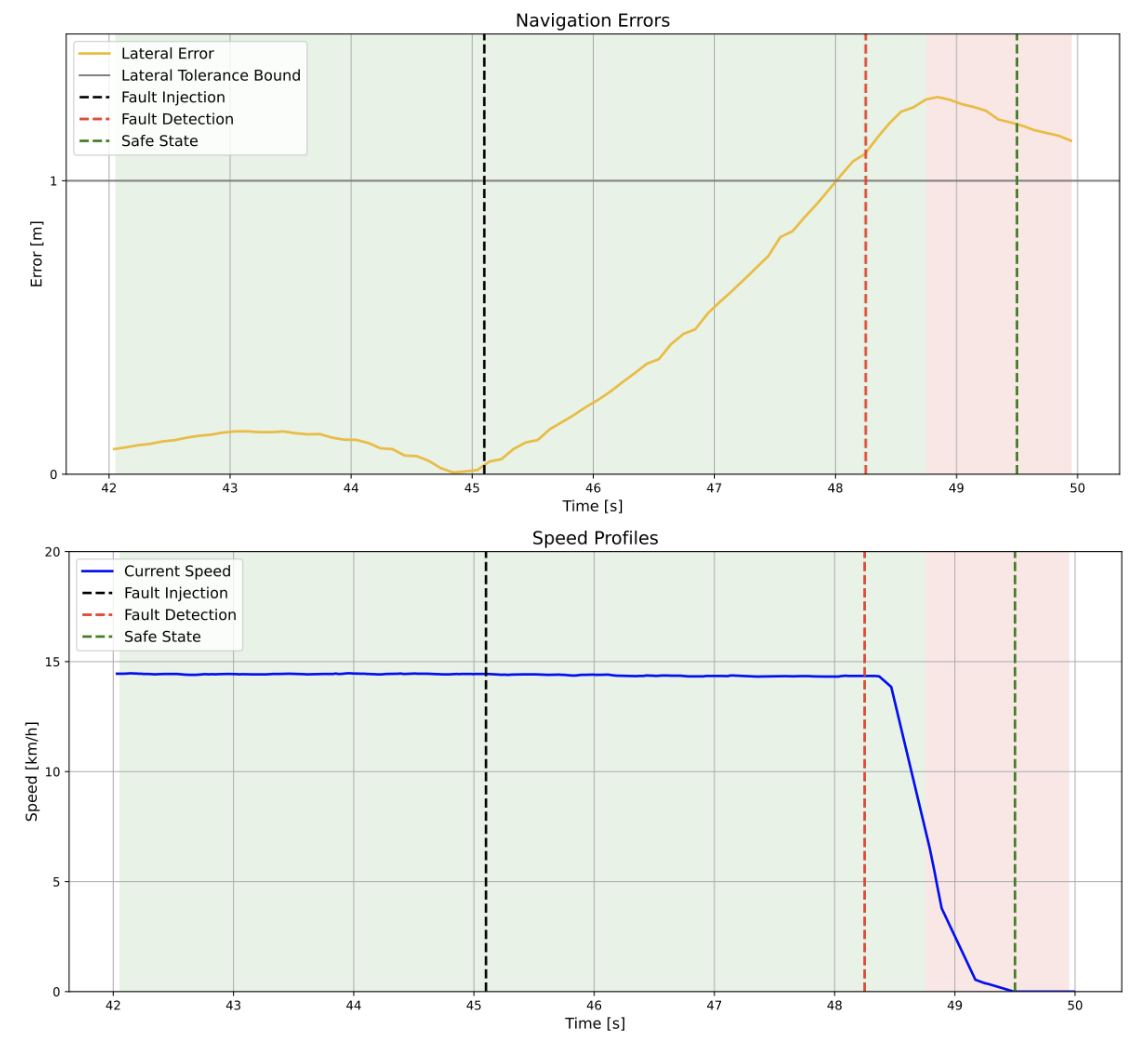}
            \caption{Experiment 21: Falsification of Event $E_{15}$, which induces HZ\_02 and characterizes the scenario where \textit{`the current vehicle position is offside the navigation reference trajectory'}. In the plots, the green areas denote periods when the autonomous mode is active, while the red areas indicate manual mode operation. The fault is introduced through a consistent steering wheel command. In the upper plot, the \ac{av}'s lateral deviation from the navigation trajectory (in absolute value) is depicted. At $t=48s$, the vehicle exceeds the safety tolerance, subsequently triggering the emergency braking system until the vehicle is fully stopped (SS\_04), as demonstrated in the lower plot.}
            \label{fig:falsification_15}
        \end{figure}

        To evaluate the efficacy of the fault recovery phase of the \ac{fusa} supervisor, it is necessary to analyze the actions taken by the \ac{av} upon fault detection. Fig.~\ref{fig:falsification_15} illustrates the behavior of the vehicle before and after the injection of the event $E_{15}$. This event is induced under operating conditions $\text{OS}_3$ (\ac{asil}~D), a scenario that generally involves interactions with other agents. Consequently, SS\_04 is activated, requiring the immediate application of the emergency brake by the supervisor once the hazard is identified.

        The results of experiment 21, as illustrated in Fig.~\ref{fig:falsification_15}, demonstrate that the time interval between the injection of an anomaly and its detection can be extended if the anomaly gradually evolves into a vehicle hazard. Consequently, the duration from when the fault is detected to when the corresponding \ac{sg} is achieved could be used to evaluate the responsiveness of the supervisor's fault recovery process. In this scenario, the \ac{av} is completely stopped 1.07~s after the fault detection.

        However, responsiveness depends on the \ac{av} and not on the \ac{fusa} supervisor architecture, since the action is invariably executed immediately after the detection of a hazard. It is important to specify within the \textit{action node} the criteria to recognize when a \ac{sg} has been achieved. In experiment 21, the emergency braking action is ended once the vehicle has come to a complete stop, 1.25~s after the hazard detection.

\section{Conclusions and Future Work} \label{sec:5}

    The \ac{bt} methodology proves to be particularly effective for the \ac{fmp}, mainly due to its inherent similarities with the \ac{fta}. Consequently, it has been shown that \ac{av} supervisors can be developed using some required information from ISO~26262 (\ac{hara} and \ac{tsc}), thus guaranteeing \ac{fusa}. An automatic conversion \ac{sw} tool, \textit{ft2bt}, which translates static into run-time safety analysis, has been developed. It avoids scalability concerns in complex systems. The tool also considers the sequential nature of \acp{bt}, which requires the strategic placement of \textit{event} and \textit{action nodes} to minimize both computational time and detection latency, and maximize reactivity to the most dangerous situations.

    The testing and falsification phases yielded favorable results in terms of the accuracy and timing of fault detection. Furthermore, these phases demonstrated the \ac{av}'s tolerance to spontaneous, albeit non-hazardous, scenario changes, attributable to meticulous parameter tuning. Regarding the fault recovery process, the \ac{bt} architecture, which links the identification of hazards with the action to mitigate them, inherently guarantees \ac{fusa} thanks to its association with static safety analysis.

    A principal advantage of the \ac{fusa} supervisor is its adaptability to various autonomous driving \ac{sw} platforms, attributed to its independence from navigation algorithms. This adaptability comes from the minimal information required from navigation inputs, such as sensors, and outputs, such as perception, estimation, and control results, to verify the accuracy of internal calculations. Consequently, this approach facilitates ongoing improvements in navigation \ac{sw} without compromising the integrity of safety run-time verifications. 
    
    However, the supervisor is designed for a specific \ac{hw} platform. Consequently, alterations in sensor configurations or modifications to actuators can directly influence the design of the \ac{fusa} supervisor. Therefore, the \textit{event} and \textit{action nodes} require redesigning, as the sources of information and the methods for altering the vehicle behavior would differ.
    
    The \ac{bt} methodology has found extensive application alongside \ac{ai} algorithms in the domains of robotics and video games. Given the increasing deployment of such algorithms in \acp{av}, the novel \ac{fusa} supervisor presents a significant advantage in ensuring safety during run-time verification. The use of descriptive \ac{ai} algorithms can facilitate the integration between neural networks and \acp{bt}.

    The study also revealed a limitation of the methodology: the supervisor is capable of detecting only one failure at a time. Consequently, in instances where multiple hazardous events occur simultaneously within the \ac{av}, only the first fault assessed is identified. An additional aspect of integrating the proposed methodology involves meticulous tuning of admissible values for event verification ($t_{anomaly}$ and tolerances), given the safety-critical nature of the tests. Finally, the proposal to find the \acp{os} could be optimized, as it is carried out in every iteration of the supervisor.

\bibliographystyle{ieeetr} 
\bibliography{main}

\begin{thebibliography}{10}

\bibitem{fagnant2015preparing}
D.~J. Fagnant and K.~Kockelman, ``Preparing a nation for autonomous vehicles: opportunities, barriers and policy recommendations,'' {\em Transportation Research Part A: Policy and Practice}, vol.~77, pp.~167--181, 2015.

\bibitem{koopman2017autonomous}
P.~Koopman and M.~Wagner, ``Autonomous vehicle safety: An interdisciplinary challenge,'' {\em IEEE Intelligent Transportation Systems Magazine}, vol.~9, no.~1, pp.~90--96, 2017.

\bibitem{iso22736}
{International Organization for Standardization}, ``{ISO/SAE PAS 22736: Taxonomy and definitions for terms related to driving automation systems for on-road motor vehicles},'' {2021}.

\bibitem{iso26262}
{International Organization for Standardization}, ``{ISO 26262: Road vehicles -- Functional safety},'' {2018}.

\bibitem{sotif}
{International Organization for Standardization}, ``{ISO 21448: Safety of the Intended Functionality (SOTIF)},'' {2022}.

\bibitem{4526677}
B.~D. Owens, M.~S. Herring, N.~Dulac, N.~G. Leveson, M.~D. Ingham, and K.~A. Weiss, ``Application of a safety-driven design methodology to an outer planet exploration mission,'' in {\em 2008 IEEE Aerospace Conference}, pp.~1--24, 2008.

\bibitem{abdulkhaleq2017using}
A.~Abdulkhaleq, S.~Wagner, D.~Lammering, H.~Boehmert, and P.~Blueher, ``Using stpa in compliance with iso 26262 for developing a safe architecture for fully automated vehicles,'' 2017.

\bibitem{schildbach2018}
G.~Schildbach, ``On the application of {ISO} 26262 in control design for automated vehicles,'' in {\em Proceedings 2nd International Workshop on Safe Control of Autonomous Vehicles, SCAV@CPSWeek 2018, Porto, Portugal, 10th April 2018} (M.~Gleirscher, S.~Kugele, and S.~Linker, eds.), vol.~269 of {\em {EPTCS}}, pp.~74--82, 2018.

\bibitem{graubohm2019}
R.~Graubohm, T.~Stolte, G.~Bagschik, M.~Steimle, and M.~Maurer, ``Functional safety concept generation within the process of preliminary design of automated driving functions at the example of an unmanned protective vehicle,'' {\em Proceedings of the Design Society: International Conference on Engineering Design}, vol.~1, no.~1, p.~2863–2872, 2019.

\bibitem{abdulkhaleq2017systematic}
A.~Abdulkhaleq, D.~Lammering, S.~Wagner, J.~Röder, N.~Balbierer, L.~Ramsauer, T.~Raste, and H.~Boehmert, ``A systematic approach based on stpa for developing a dependable architecture for fully automated driving vehicles,'' {\em Procedia Engineering}, vol.~179, pp.~41--51, 2017.
\newblock 4th European STAMP Workshop 2016, ESW 2016, 13-15 September 2016, Zurich, Switzerland.

\bibitem{8479057}
M.~A. Gosavi, B.~B. Rhoades, and J.~M. Conrad, ``Application of functional safety in autonomous vehicles using iso 26262 standard: A survey,'' in {\em SoutheastCon 2018}, pp.~1--6, 2018.

\bibitem{koopman2019}
P.~Koopman, U.~Ferrell, F.~Fratrik, and M.~Wagner, ``A safety standard approach for fully autonomous vehicles,'' in {\em Computer Safety, Reliability, and Security: SAFECOMP 2019 Workshops, ASSURE, DECSoS, SASSUR, STRIVE, and WAISE, Turku, Finland, September 10, 2019, Proceedings}, (Berlin, Heidelberg), p.~326–332, Springer-Verlag, 2019.

\bibitem{macher2017}
G.~Macher, A.~Much, A.~Riel, R.~Messnarz, and C.~Kreiner, ``Automotive spice, safety and cybersecurity integration,'' in {\em Computer Safety, Reliability, and Security} (S.~Tonetta, E.~Schoitsch, and F.~Bitsch, eds.), (Cham), pp.~273--285, Springer International Publishing, 2017.

\bibitem{ulbrich2017functional}
S.~Ulbrich, A.~Reschka, J.~Rieken, S.~Ernst, G.~Bagschik, F.~Dierkes, M.~Nolte, and M.~Maurer, ``Towards a functional system architecture for automated vehicles,'' 2017.

\bibitem{bagschik2018system}
G.~Bagschik, M.~Nolte, S.~Ernst, and M.~Maurer, ``A system's perspective towards an architecture framework for safe automated vehicles,'' in {\em 2018 21st International Conference on Intelligent Transportation Systems (ITSC)}, pp.~2438--2445, IEEE, 2018.

\bibitem{tabani2019}
H.~Tabani, L.~Kosmidis, J.~Abella, F.~J. Cazorla, and G.~Bernat, ``Assessing the adherence of an industrial autonomous driving framework to iso 26262 software guidelines,'' in {\em 2019 56th ACM/IEEE Design Automation Conference (DAC)}, pp.~1--6, 2019.

\bibitem{8772139}
Y.~Wang, Z.~Liu, Z.~Zuo, Z.~Li, L.~Wang, and X.~Luo, ``Trajectory planning and safety assessment of autonomous vehicles based on motion prediction and model predictive control,'' {\em IEEE Transactions on Vehicular Technology}, vol.~68, no.~9, pp.~8546--8556, 2019.

\bibitem{ifqir2022fault}
S.~Ifqir, C.~Combastel, A.~Zolghadri, G.~Alcalay, P.~Goupil, and S.~Merlet, ``Fault tolerant multi-sensor data fusion for autonomous navigation in future civil aviation operations,'' {\em Control Engineering Practice}, vol.~123, p.~105132, 2022.

\bibitem{prieto2020localization}
D.~Prieto~Francia, ``Autonomous vehicle localization using state estimation techniques,'' Master's thesis, Universitat Polit{\`e}cnica de Catalunya, 2020.

\bibitem{vicens2022localization}
J.-M. Vicens~Sancho, ``Design of a state and localization estimator for an autonomous vehicle,'' Master's thesis, Universitat Polit{\`e}cnica de Catalunya, 2022.

\bibitem{blanke2006}
M.~Blanke, M.~Kinnaert, J.~Lunze, M.~Staroswiecki, and J.~Schröder, {\em Diagnosis and Fault-Tolerant Control}, ch.~1, pp.~1--45.
\newblock Springer International Publishing, 01 2006.

\bibitem{lemmon1999supervisory}
M.~Lemmon, K.~He, and I.~Markovsky, ``Supervisory hybrid systems,'' {\em IEEE Control Systems Magazine}, vol.~19, no.~4, pp.~42--55, 1999.

\bibitem{vento2013hybrid}
J.~Vento~Maldonado, L.~Trav{\'e}-Massuy{\`e}s, R.~Sarrate~Estruch, and V.~Puig~Cayuela, ``Hybrid automaton incremental construction for online diagnosis,'' in {\em Proceedings DX'13}, pp.~186--191, 2013.

\bibitem{heffernan2014runtime}
D.~Heffernan, C.~MacNamee, and P.~Fogarty, ``Runtime verification monitoring for automotive embedded systems using the iso 26262 functional safety standard as a guide for the definition of the monitored properties,'' {\em IET Software}, vol.~8, no.~5, pp.~193--203, 2014.

\bibitem{kaalen2019}
S.~Kaalen, M.~Nyberg, and C.~Bondesson, ``Tool-supported dependability analysis of semi-markov processes with application to autonomous driving,'' in {\em 2019 4th International Conference on System Reliability and Safety (ICSRS)}, pp.~126--135, 2019.

\bibitem{dorff2020}
S.~v. Dorff, B.~Böddeker, M.~Kneissl, and M.~Fränzle, ``A fail-safe architecture for automated driving,'' in {\em 2020 Design, Automation \& Test in Europe Conference \& Exhibition (DATE)}, pp.~828--833, 2020.

\bibitem{stolte2020towards}
T.~Stolte, R.~Graubohm, I.~Jatzkowski, M.~Maurer, S.~Ackermann, B.~Klamann, M.~Lippert, and H.~Winner, ``Towards safety concepts for automated vehicles by the example of the project unicaragil,'' in {\em Aachen Colloquium Sustainable Mobility 2020}, 10 2020.

\bibitem{CUER201829}
R.~Cuer, L.~Piétrac, E.~Niel, S.~Diallo, N.~Minoiu-Enache, and C.~Dang-Van-Nhan, ``A formal framework for the safe design of the autonomous driving supervision,'' {\em Reliability Engineering \& System Safety}, vol.~174, pp.~29--40, 2018.

\bibitem{isla2008halo}
D.~Isla, ``Halo 3 - building a better battle,'' in {\em Game Developers Conference}, 2008.

\bibitem{colledanchise2018behavior}
M.~Colledanchise and P.~Ögren, ``Behavior trees in robotics and ai,'' July 2018.

\bibitem{iovino2022survey}
M.~Iovino, E.~Scukins, J.~Styrud, P.~Ögren, and C.~Smith, ``A survey of behavior trees in robotics and ai,'' {\em Robotics and Autonomous Systems}, vol.~154, p.~104096, 2022.

\bibitem{marzinotto2014towards}
A.~Marzinotto, M.~Colledanchise, C.~Smith, and P.~Ögren, ``Towards a unified behavior trees framework for robot control,'' in {\em 2014 IEEE International Conference on Robotics and Automation (ICRA)}, pp.~5420--5427, 2014.

\bibitem{colledanchise2017how}
M.~Colledanchise and P.~Ögren, ``How behavior trees modularize hybrid control systems and generalize sequential behavior compositions, the subsumption architecture, and decision trees,'' {\em IEEE Transactions on Robotics}, vol.~33, no.~2, pp.~372--389, 2017.

\bibitem{ogren2012increasing}
P.~Ögren, ``Increasing modularity of uav control systems using computer game behavior trees,'' in {\em Aiaa guidance, navigation, and control conference}, p.~4458, 2012.

\bibitem{palma2011extending}
R.~Palma, P.~González-Calero, M.~Gómez-Martín, and P.~Gómez-Martín, ``Extending case-based planning with behavior trees,'' in {\em Proceedings of the 24th International Florida Artificial Intelligence Research Society, FLAIRS - 24}, pp.~407--412, 2011.

\bibitem{bagnell2012integrated}
J.~A. Bagnell, F.~Cavalcanti, L.~Cui, T.~Galluzzo, M.~Hebert, M.~Kazemi, M.~Klingensmith, J.~Libby, T.~Y. Liu, N.~Pollard, M.~Pivtoraiko, J.-S. Valois, and R.~Zhu, ``An integrated system for autonomous robotics manipulation,'' in {\em 2012 IEEE/RSJ International Conference on Intelligent Robots and Systems}, pp.~2955--2962, 2012.

\bibitem{olsson2016behavior}
M.~Olsson, ``Behavior trees for decision-making in autonomous driving,'' Master's thesis, KTH, School of Computer Science and Communication (CSC), 2016.

\bibitem{rovida2018motion}
F.~Rovida, D.~Wuthier, B.~Grossmann, M.~Fumagalli, and V.~Krüger, ``Motion generators combined with behavior trees: A novel approach to skill modelling,'' in {\em 2018 IEEE/RSJ International Conference on Intelligent Robots and Systems (IROS)}, pp.~5964--5971, 2018.

\bibitem{paxton2019representing}
C.~Paxton, N.~Ratliff, C.~Eppner, and D.~Fox, ``Representing robot task plans as robust logical-dynamical systems,'' in {\em 2019 IEEE/RSJ International Conference on Intelligent Robots and Systems (IROS)}, pp.~5588--5595, 2019.

\bibitem{faconti2018behaviortree}
D.~Faconti and M.~Colledanchise, ``Behaviortree.cpp library documentation,'' 2018.

\bibitem{tadewos2019decentralized}
T.~G. Tadewos, L.~Shamgah, and A.~Karimoddini, ``On-the-fly decentralized tasking of autonomous vehicles,'' in {\em 2019 IEEE 58th Conference on Decision and Control (CDC)}, pp.~2770--2775, 2019.

\bibitem{ghzouli2020behavior}
R.~Ghzouli, T.~Berger, E.~B. Johnsen, S.~Dragule, and A.~Wasowski, ``Behavior trees in action: A study of robotics applications,'' in {\em Proceedings of the 13th ACM SIGPLAN International Conference on Software Language Engineering}, SLE 2020, (New York, NY, USA), p.~196–209, Association for Computing Machinery, 2020.

\bibitem{lindsay2010safetyassessment}
P.~A. Lindsay, K.~Winter, and N.~Yatapanage, ``Safety assessment using behavior trees and model checking,'' in {\em 2010 8th IEEE International Conference on Software Engineering and Formal Methods}, pp.~181--190, 2010.

\bibitem{Survey_Safety}
S.~Riedmaier, T.~Ponn, D.~Ludwig, B.~Schick, and F.~Diermeyer, ``Survey on scenario-based safety assessment of automated vehicles,'' {\em IEEE Access}, vol.~8, pp.~87456--87477, 2020.

\bibitem{koschi2019computationally}
M.~Koschi, C.~Pek, S.~Maierhofer, and M.~Althoff, ``Computationally efficient safety falsification of adaptive cruise control systems,'' in {\em 2019 IEEE Intelligent Transportation Systems Conference (ITSC)}, pp.~2879--2886, IEEE, 2019.

\bibitem{shalevshwartz2018formal}
S.~Shalev-Shwartz, S.~Shammah, and A.~Shashua, ``On a formal model of safe and scalable self-driving cars,'' 2018.

\bibitem{winkle2016safety}
T.~Winkle, ``Safety benefits of automated vehicles: Extended findings from accident research for development, validation and testing,'' in {\em Autonomous driving}, pp.~335--364, Springer, 2016.

\bibitem{vesely1981fault}
W.~E. Vesely, F.~F. Goldberg, N.~H. Roberts, D.~F. Haasl, {\em et~al.}, {\em Fault tree handbook}.
\newblock Systems and Reliability Research, Office of Nuclear Regulatory Research, US~…, 1981.

\bibitem{schonemann2019fault}
V.~Schönemann, H.~Winner, T.~Glock, E.~Sax, B.~Boeddeker, S.~vom Dorff, G.~Verhaeg, F.~Tronci, and G.~G. Padilla, ``Fault tree-based derivation of safety requirements for automated driving on the example of cooperative valet parking,'' in {\em 26th International Technical Conference on the Enhanced Safety of Vehicles (ESV) 2019}, vol.~9, 2019.

\bibitem{dugan1992dynamic}
J.~Dugan, S.~Bavuso, and M.~Boyd, ``Dynamic fault-tree models for fault-tolerant computer systems,'' {\em IEEE Transactions on Reliability}, vol.~41, no.~3, pp.~363--377, 1992.

\bibitem{volk2018fast}
M.~Volk, S.~Junges, and J.-P. Katoen, ``Fast dynamic fault tree analysis by model checking techniques,'' {\em IEEE Transactions on Industrial Informatics}, vol.~14, no.~1, pp.~370--379, 2018.

\bibitem{kabir2019conceptual}
S.~Kabir, K.~Aslansefat, I.~Sorokos, Y.~Papadopoulos, and Y.~Gheraibia, ``A conceptual framework to incorporate complex basic events in hip-hops,'' in {\em Model-Based Safety and Assessment} (Y.~Papadopoulos, K.~Aslansefat, P.~Katsaros, and M.~Bozzano, eds.), (Cham), pp.~109--124, Springer International Publishing, 2019.

\end{thebibliography}

\section*{Biographies}

\begin{IEEEbiography}[{\includegraphics[width=1in,height=1.25in,clip,keepaspectratio]{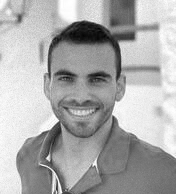}}]{Carlos Conejo}
received the B.Sc. and M.Sc. degrees in industrial technology engineering, specializing in automatic control and robotics, from Universitat Politècnica de Catalunya–BarcelonaTech (UPC) in 2018 and 2021, respectively. He is a Ph.D. student at the Institut de Robòtica i Informàtica Industrial (IRI), CSIC-UPC, collaborating with the Renault Group. He is also an associate professor in the Department of Automatic Control. His research interests include fault diagnosis, fault-tolerant control, and functional safety for autonomous vehicles.
\end{IEEEbiography}

\begin{IEEEbiography}[{\includegraphics[width=1in,height=1.25in,clip,keepaspectratio]{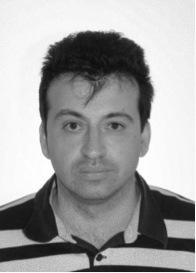}}]{Vicenç Puig}
received the B.Sc./M.Sc. degree in telecommunications engineering and the
Ph.D. degree in automatic control, vision, and robotics from Universitat Politècnica de Catalunya–BarcelonaTech (UPC), in 1993 and 1999, respectively. He is currently a Full Professor with the Automatic Control Department, UPC, and a Researcher with the Institut de Robòtica
i Informàtica Industrial (IRI), CSIC-UPC. He is the Director of the Automatic Control Department and the Head of the Research Group on Advanced Control Systems (SAC), UPC. He has developed important scientific contributions in the areas of fault diagnosis and fault tolerant control, using interval and linear-parameter-varying models using set-based approaches.
He is also the Chair of the IFAC Safeprocess TC Committee 6.4. He was
the General Chair of the 3rd IEEE Conference on Control and Fault-Tolerant
Systems (SysTol 2016) and the IPC Chair of IFAC Safeprocess 2018.
\end{IEEEbiography}

\begin{IEEEbiography}[{\includegraphics[width=1in,height=1.25in,clip,keepaspectratio]{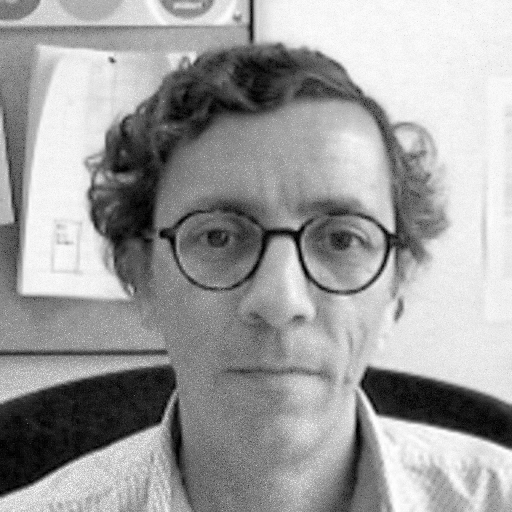}}]{Bernardo Morcego}
is an Associate Professor at the Universitat Politecnica de Catalunya (UPC). He received a PhD degree in
Computer Science from the UPC in 2000. He has been teaching several subjects in the area of automatic control in the schools of Engineering
and Aeronautics in Terrassa and Barcelona. He is a member of the Research Center for Supervision, Safety and Automatic Control of
UPC. His research interests include UAV control systems, intelligent control and computer vision applications.
\end{IEEEbiography}

\begin{IEEEbiography}[{\includegraphics[width=1in,height=1.25in,clip,keepaspectratio]{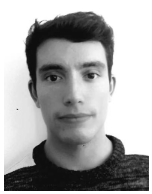}}]{Francisco Navas}
received the Ph.D. degree from the RITS Team, INRIA, Paris, and MINES ParisTech, PSL Research University, in November 2018. Since 2018, he has been with the Research Department, AKKA Technologies. His research interests include autonomous vehicles, Youla-Kucera parameterization, switching controllers, and vehicle-infrastructure cooperation.
\end{IEEEbiography}

\begin{IEEEbiography}[{\includegraphics[width=1in,height=1.25in,clip,keepaspectratio]{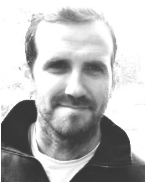}}]{Vicente Milanés}
received the Ph.D. degree in electronic engineering from the University of Alcalá, Madrid, Spain, in 2010. He was with the AUTOPIA Program with the Center for Automation and Robotics (UPM-CSIC), Spain, from 2006 to 2011. In 2014, he joined the RITS Team, INRIA, France. Since 2016, he has been with the Research Department, Renault, France. His research interests include autonomous vehicles, vehicle dynamic control, intelligent traffic and transport infrastructures, and vehicle-infrastructure cooperation. He was awarded with a two-year Fulbright Fellowship at California PATH, UC Berkeley. He has been awarded with the Best Paper Award in three conferences and his Ph.D. has received three major awards.
\end{IEEEbiography}

\end{document}